\documentclass[runningheads]{llncs}
\usepackage{graphicx}
\usepackage{tikz}
\usepackage{comment}
\usepackage{amsmath,amssymb} 
\usepackage{color}

\usepackage{booktabs}
\usepackage{multirow}
\usepackage{makecell}
\usepackage{pifont}
\usepackage[pagebackref=true,breaklinks=true,colorlinks,citecolor=green,linkcolor=blue,bookmarks=false]{hyperref}

\usepackage{array}
\usepackage{caption}
\usepackage{floatrow}
\usepackage{tabularx}

\newcommand{\cmark}{\ding{51}}%
\newcommand{\xmark}{\ding{55}}%
\newcommand{\calM}{{\mathcal{M}}}

\newcommand{\calS}{{\mathcal{S}}}

\newcommand{\be}{\begin{eqnarray}}
\newcommand{\ee}{\end{eqnarray}}
\newcommand{\bee}{\begin{eqnarray*}}
\newcommand{\eee}{\end{eqnarray*}}

\newcommand{\matrixb}{\left[ \begin{array}}
\newcommand{\matrixe}{\end{array} \right]}   

\newcommand{\Tref}[1]{Table~\ref{#1}}

\newcommand{\Fref}[1]{Fig.~\ref{#1}}

\newcommand{\Sref}[1]{Sec.~\ref{#1}}

\def\eg{\emph{e.g.}}

\def\etal{\emph{et al.}}
\def\ie{\emph{i.e.}}

\usepackage{colortbl}
\definecolor{steelblue}{rgb}{0.27, 0.51, 0.7}

\newcommand{\mytabular}[1]{\centering\scalebox{0.75}{#1}}

\begin{document}
\pagestyle{headings}
\mainmatter
\def\ECCVSubNumber{4736}  

\title{The Anatomy of Video Editing: A Dataset and Benchmark Suite for AI-Assisted Video Editing} 

\titlerunning{ECCV-22 submission ID \ECCVSubNumber} 
\authorrunning{ECCV-22 submission ID \ECCVSubNumber} 
\author{Anonymous ECCV submission}
\institute{Paper ID \ECCVSubNumber}

\titlerunning{The Anatomy of Video Editing}
%
\author{Dawit Mureja Argaw\inst{1,2} \and
Fabian Caba Heilbron\inst{1} \and
Joon-Young Lee\inst{1} \and \\
Markus Woodson\inst{1} \and
In So Kweon\inst{2}}

\authorrunning{D. Mureja et al.}
%
\institute{Adobe Research \and KAIST
}
\maketitle

\begin{abstract}
Machine learning is transforming the video editing industry. Recent advances in computer vision have leveled-up video editing tasks such as intelligent reframing, rotoscoping, color grading, or applying digital makeups. However, most of the solutions have focused on video manipulation and VFX. This work introduces the Anatomy of Video Editing, a dataset, and benchmark, to foster research in AI-assisted video editing. Our benchmark suite focuses on video editing tasks, beyond visual effects, such as automatic footage organization and assisted video assembling. To enable research on these fronts, we annotate more than 1.5M tags, with relevant concepts to cinematography, from 196176 shots sampled from movie scenes. We establish competitive baseline methods and detailed analyses for each of the tasks. We hope our work sparks innovative research towards underexplored areas of AI-assisted video editing. Code is available at: \href{https://github.com/dawitmureja/AVE.git}{https://github.com/dawitmureja/AVE.git}.

\end{abstract}
\section{Introduction}
\vspace{-2mm}
What does the future of video editing look like? Arguably AI-based technologies will have a strong influence on this creative industry. In fact, the computer vision community has already delivered technologies such as automatic rotoscoping \cite{oh2019video} as a teaser of the opportunities to transform video editing. Most research development has centered around enabling AI-based VFX (visual effects) \cite{oh2019video,liu2021learning,kasten2021layered,gao2020flow,lu2021omnimatte,patwardhan2007video,smith2017harnessing}; however, editing video involves more than that. Despite their importance towards AI-assisted video editing, topics such as understanding cinematography concepts for automatic organization and assisting editors to assembly edits remain underexplored in the computer vision community.

Research progress toward AI-assisted video editing has been hindered by the lack of formally defined tasks relevant to the editing process. This observation sparked recent works to study film properties or learn cutting patterns from movie data \cite{huang2020movienet,wu2021towards,pardo2021learning}. MovieNet \cite{huang2020movienet} touches on cinematography style by providing annotations for two attributes: view scale and camera movement. While automatically tagging these concepts already provides value to the automatic organization, the cinematographic vocabulary comprises a much larger set. Learning to Cut \cite{pardo2021learning} recommends the best moments to cut a pair of shots by looking at motion. This task is indeed an important editing task, but there are still other decisions such as establishing the order of shots and the most suitable composition to assist video assembling. The progress is exciting, but the existing tasks cover a limited span of video editing.

\begin{figure*}[!t]
    \centering
    \includegraphics[width=1.0\linewidth]{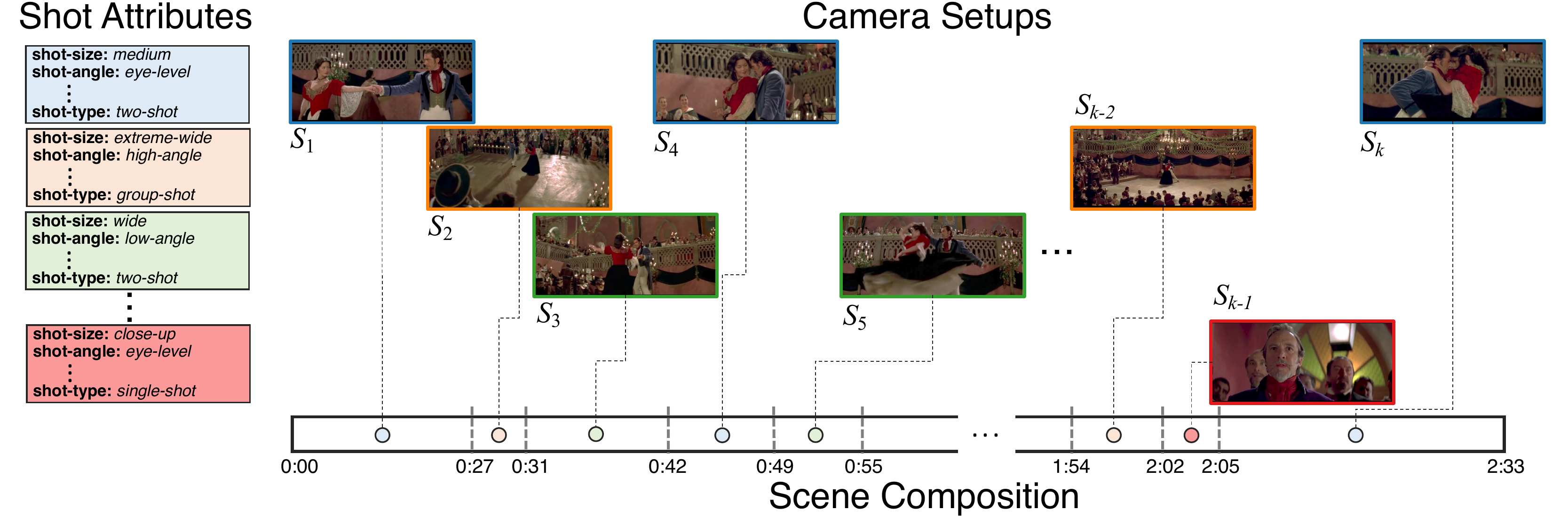}
    \caption{\small Anatomy of Video Editing (\texttt{AVE}) dataset and benchmark suite. Our dataset decomposes movie scenes into sequence of shots by identifying the cuts in the scene. Each shot composing the scene has cinematographic attributes and camera setup labels.}
    \label{fig:pull-figure}
    \vspace{-2mm}
\end{figure*}

To enable research development on AI-assisted video editing, we introduce the Anatomy of Video Editing (\texttt{AVE}), a dataset and benchmark suite. Movies require extensive hours of assistant editors to organize and tag footage and they depict the most creative and artistic forms of video editing. These properties motivate us to build the \texttt{AVE} dataset upon 5591 movie scenes. We recover the temporal composition of each scene by annotating the shot transitions and the camera setups. In total, we annotate 196176 shots among eight cinematography properties, yielding more than 1.5M labels. Fig.~\ref{fig:pull-figure} illustrates one annotated movie scene from our dataset.

Our benchmark suite facilitates research in two areas to advance AI-assisted video editing. We define two tasks related to automatically organizing footage and introduce three tasks that aim to learn editors’ patterns in video assembling. Equally crucial to defining the right tasks is establishing solid baselines, metrics, and initial analyses. Our baselines include modern video understanding methods, providing a competitive start. Nevertheless, our analyses discuss opportunities to develop new models to improve upon our baselines on the proposed tasks.

\noindent\textbf{Contributions.} To summarize, our contributions are two-fold:\\
\noindent \textbf{(1)} We introduce the \texttt{AVE} dataset, which includes the composition of 5591 movie scenes with more than 1.5M cinematography labels for 196176 shots (Sec. \ref{sec:dataset}). \\
\noindent \textbf{(2)} We establish a benchmark suite that includes five different tasks for AI-assisted video editing (Sec. \ref{sec:benchmark}). Along with each task definition, we implement competitive baselines and provide extensive experimental analyses (Sec. \ref{sec:experiment}).

\vspace{-3mm}
\section{Related Works}
\vspace{-2mm}
\subsubsection{Movie Datasets.}
Several movie-based datasets have been presented by previous works~\cite{bain2020condensed,zhu2015aligning,tapaswi2016movieqa,liu2021multi,wu2021towards,huang2020movienet} for various video understanding tasks. Zhu \etal~\cite{zhu2015aligning} proposed the \texttt{MovieBook} dataset for aligning stories from books and movies. Tapaswi \etal~\cite{tapaswi2016movieqa} introduced the \texttt{MovieQA} dataset. Bain \etal~\cite{bain2020condensed} proposed \texttt{Condensed Movies}, which contains 33K short movie clips and high-level text descriptions for text-to-video retrieval task. Liu \etal~\cite{liu2021multi} proposed \texttt{MUSES}, which contains 31477 clips collected from several drama episodes and annotated with 25 action categories for temporal event localization task. Wu \etal~\cite{wu2021towards} explored the long-term video understanding problem on movie clips dataset by formulating high-level prediction tasks. Recently, Huang \etal~\cite{huang2020movienet} proposed the \texttt{MovieNet}, which contains 1100 full movies and 60K trailers with diverse annotations to learn various tasks such as genre prediction, scene segmentation, and character recognition. While previous works primarily focus on high-level video content understanding tasks at a clip level, our work goes one step further and studies the sequence of shots in movie scenes by proposing a large-scale dataset with shot-level attributes and scene-level composition annotations.
\vspace{-4mm}
\subsubsection{Shot Assembly in Video Editing.} Here we focus on previous works that studied cinematography patterns in film editing. For an in-depth discussion on video editing, in general, we invite the reader to \cite{zhang2022ai}. Given a script and different takes of a scene, Leake \etal~\cite{leake2017computational} attempted to generate a scene by selecting a relevant clip for each line of dialogue in a given script. Wu \etal~\cite{wu2018thinking,wu2016analysing} used shot relation attributes to formulate film editing pattern syntax and provided an interactive editing platform. Minh \etal~\cite{hoai2014thread} decomposed a video scene into an ordered sequence of relevant shots to remove abrupt discontinuities for improved human action recognition. Other works~\cite{baxter2013comparing,rao2020unified,pardo2021learning,pardo2021moviecuts} studied cutting patterns in movie scenes. Although some previous works~\cite{wu2016analysing,wu2018thinking,leake2017computational} attempted to study the film editing process, they are limited to a particular type of scene, \eg~dialogue, or cannot generalize to new editing patterns beyond a predefined set. Our work aims at learning general editing patterns from a large set of publicly-available movie scenes using a data-driven approach. 
\vspace{-3mm}
\section{Anatomy of Video Editing: Dataset}
\label{sec:dataset}
\vspace{-2mm}
Here we describe the collection of the Anatomy of Video Editing (\texttt{AVE}) dataset, a large-scale shot attribute set that contains approximately $196,176$ shots dissected from $5,591$ publicly available movie scenes \cite{bain2020condensed,pardo2021learning}\footnote{We crawled the movie scenes from the \href{https://www.youtube.com/user/movieclips}{MovieClips YouTube Channel}.}.
\vspace{-3mm}
\subsection{Shot Attributes}
\label{sec:shot_attributes}
Following the standard definition of shot properties in cinematography\cite{metz1991film,dancyger2018technique}, we label eight attributes, which we define below, for each shot in \texttt{AVE}.

\textbf{Shot size} is defined as how much of the setting or subject is displayed within a given shot. Shot size has \textit{five} categories: 1) \texttt{Extreme wide} (\texttt{EW}) shots barely show the subject and the shot's main focus is the subject's surrounding; 2) \texttt{Wide} (\texttt{W}) shots, also known as long shot, show the entire subject and their relation to the surrounding environment; 3) \texttt{Medium} (\texttt{M}) shots depict the subject approximately from the waist up emphasizing both the subject and their surrounding; 4) \texttt{Close-up} (\texttt{CU}) shots are taken at a close range intended to show greater detail to the viewer; 5) \texttt{Extreme close-up} (\texttt{ECU}) shots frame a subject very closely where the outer portions of the subject are often cut off by the frame's edges.

\textbf{Shot angle} is the location where the camera is placed to take a shot. Shot angle has \textit{five} categories: 1) \texttt{Aerial} (\texttt{A}) shot is captured from an elevated vantage point; 2) \texttt{Overhead} (\texttt{O}) shot is when the camera is placed directly above the subject; 3) \texttt{Eye level} (\texttt{EL}) shot is a shot where the camera is positioned directly at the subject's eye level; 4) \texttt{High angle} (\texttt{HA}) shot is when the camera points down on the subject from above; 5) \texttt{Low angle} (\texttt{LA}) shot is when the camera is positioned below the eye level and looks up at the subject.

\textbf{Shot type} refers to the composition of a shot in terms of the number of featured subjects and their physical relationship to each other and the camera. Shot type has \textit{six} categories: 1) \texttt{Over-the-shoulder} (\texttt{OTS}) shot shows the main subject from behind the shoulder of another subject; 2) \texttt{Single} (\texttt{S}) shot captures one subject; 3) \texttt{Two} (\texttt{2}) shot has two subjects featured in the frame; 4) \texttt{Three} (\texttt{3}) shot has three characters in the frame 5) \texttt{Insert} (\texttt{I}) shot is any shot whose purpose is to draw the viewer’s attention to a specific detail within a scene; 6) \texttt{Group} (\texttt{G}) shot features a group of subjects in the shot.

\textbf{Shot motion} is defined as the movement of the camera when taking a shot. Shot motion has \textit{five} categories: 1) \texttt{Pan/Truck} (\texttt{P/T}) shot is when the camera is moving horizontally while its base remains in a fixed position; 2) \texttt{Tilt/Pedestal} (\texttt{T/P}) shot is when the camera moves vertically up or down with its base fixated to a certain point; 3) \texttt{Locked} (\texttt{L}) shot is taken without shifting the position of the camera; 4) \texttt{Zoom/Dolly} (\texttt{Z/D}) shot is when the camera moves forward and backward adding depth to a scene; 5) \texttt{Handheld} (\texttt{H}) shot is taken with the camera being supported only by the operator's hands and shoulder.

\textbf{Shot location} refers to the environment where the shot is taken. Shot location has \textit{two} categories: 1) \texttt{Exterior} (\texttt{Ext}) shot is taken outdoors; 2) \texttt{Interior} (\texttt{Int}) shot is taken indoors.

\textbf{Shot subject} is the main subject featured or conveyed in the shot. Shot subject has \textit{seven} categories: 1) \texttt{Animal}, 2) \texttt{Location}, 3) \texttt{Object}, 4) \texttt{Human}, 5) \texttt{Limb}, 6) \texttt{Face} and 7) \texttt{Text}.

\textbf{Num. of people} is the number of humans displayed in the shot and it has \textit{six} categories: 1) \texttt{None} (\texttt{0}), 2) \texttt{One} (\texttt{1}), 3) \texttt{Two} (\texttt{2}), 4) \texttt{Three} (\texttt{3}), 5) \texttt{Four} (\texttt{4}) and 6) \texttt{Five} (\texttt{5}) if the shot has five or more people.

\textbf{Sound source} refers to the source of sound in the shot. Sound source has \textit{four} categories: 1) \texttt{On screen} (\texttt{OnS}) - the source is a subject within the shot; 2) \texttt{Off screen} (\texttt{OfS}) - the sound comes from a subject not shown in the shot; 3) \texttt{External narration} (\texttt{EN}) - the source is a narration outside the shot; 4) \texttt{External music} (\texttt{EM}) - the only sound in the shot is music.
\vspace{-3mm}
\subsection{Scene Composition and Camera Setups}
In addition to the attributes of the individual shots, we also provide annotation for the shot sequence composition, where we label the \texttt{start} and \texttt{end} time of each shot within a scene. We also group the shots that belong to the same \texttt{camera setup} and annotate the total number of takes used in the edited scene. These annotations will enable our studies on shot pattern selection and sequencing.
\vspace{-3mm}
\subsection{Annotation Procedure}
We recruited a task force of 15 professional video editors. To reduce the amount of manual effort, we pre-segmented each video (scene) into shots using a pre-trained shot-boundary detector \cite{souvcek2020transnet}. Then, the annotation process consisted of two steps. First, we asked the annotators to verify the automatic shot boundaries are correct and to group the shots in a scene by camera setup. Second, we asked the task force to label each shot with the attributes listed in Sec. \ref{sec:shot_attributes}. 
\vspace{-3mm}
\subsection{Dataset Statistics}
\vspace{-1mm}
The Anatomy of Video Editing  (\texttt{AVE}) dataset consists of 196,176 holistically annotated shots, collected from 5,591 movie scenes that cover a wide range of genres. In \Tref{tbl:data_compare}, we compare \texttt{AVE} with related datasets. Our dataset is considerably larger in size with significantly more comprehensive, and relevant to video editing, annotations. Previous works primarily focus on individual shot properties to analyze certain cinematic techniques, but \texttt{AVE} goes beyond shot level attributes by offering the temporal sequencing of shots and the composition of scenes. \Tref{tbl:data_stat} presents detailed statistics for our \texttt{train-val-test} splits. 

\begin{table}[!t]
\setlength{\tabcolsep}{5pt}
\begin{center}
\caption{\small Comparison with related datasets.}
\vspace{-2.5mm}
\label{tbl:data_compare}
\mytabular{
\begin{tabular}{l|ccccc}
\toprule
 & \thead{Num. of \\ Shots} & \thead{Num. of \\ Videos} & \thead{Num. of \\ Shot Attributes} & \thead{Scene \\ Composition} & \thead{Camera \\ Setup} \\
\midrule
Lie 2014 \cite{bhattacharya2014classification} & 327 & 327 & 1 & \xmark & \xmark \\
Context 2011 \cite{xu2011using} & 3,206 & 4 & 1 & \xmark & \xmark \\
Cinema 2013 \cite{canini2013classifying} & 3,000 & 12 & 1 & \xmark & \xmark \\
Taxon 2009 \cite{wang2009taxonomy} & 5054 & 7 & 1 & \xmark & \xmark \\
MovieSD 2020 \cite{rao2020unified} & 46,857 & 7,858 & 2 & \xmark & \xmark \\
\texttt{AVE} & 196,176 & 5,591 & 8 & \cmark & \cmark \\
\bottomrule
\end{tabular}
}
\end{center}
\vspace{-3mm}
\end{table}

\begin{table}[!t]
\setlength{\tabcolsep}{6pt}
\begin{center}
\caption{\small Statistics of \texttt{AVE}.}
\vspace{-2.5mm}
\label{tbl:data_stat}
\mytabular{
\begin{tabular}{l|cccc}
\toprule
 & Train & Val & Test & Total \\
\midrule
Num. of scenes &  3914 & 559 &  1,118 & 5,591 \\
Num. of shots & 151,053 & 15,040  & 30,083  & 196,176\\
Avg. duration of shots (sec.) & 3.83 & 3.71 & 3.78 & 3.81 \\
Avg. number of shots per scene & 34.71  & 35.42 & 35.69 & 35.09 \\
Avg. number of camera setups & 5.71 & 6.11 & 5.69 & 5.74 \\
\bottomrule
\end{tabular}
}
\end{center}
\vspace{-3mm}
\end{table}
\vspace{-3mm}
\section{Anatomy of Video Editing: Benchmark Suite}
\label{sec:benchmark}
\vspace{-2mm}
In this section, we introduce five tasks for AI-assisted Video editing. The first two tasks focus on benchmarking the ability to automatically organize and tag footage according to cinematography properties. The last three tasks center around predicting editing patterns used in movie scenes. 

\vspace{-4mm}
\subsubsection{Notation.} Let $\calM$ represent a movie scene which is defined as a sequence of $k$ shots, \ie~$\calM = \{\calS_1, \calS_2, \ldots, \calS_k\}$, where $\calS$ denotes a shot clip. Each shot $\calS$ is composed of visual ($v$) and audio ($a$) representations, \ie~$\calS = \langle v,a\rangle$. The audio-visual features encoded from each shot clips are denoted as $\{u_1,u_2, \ldots, u_k\}$.

\begin{figure*}[!t]
    \centering
    \includegraphics[width=1.0\linewidth,trim={11.25cm 16.5cm 25cm 8.9cm},clip]{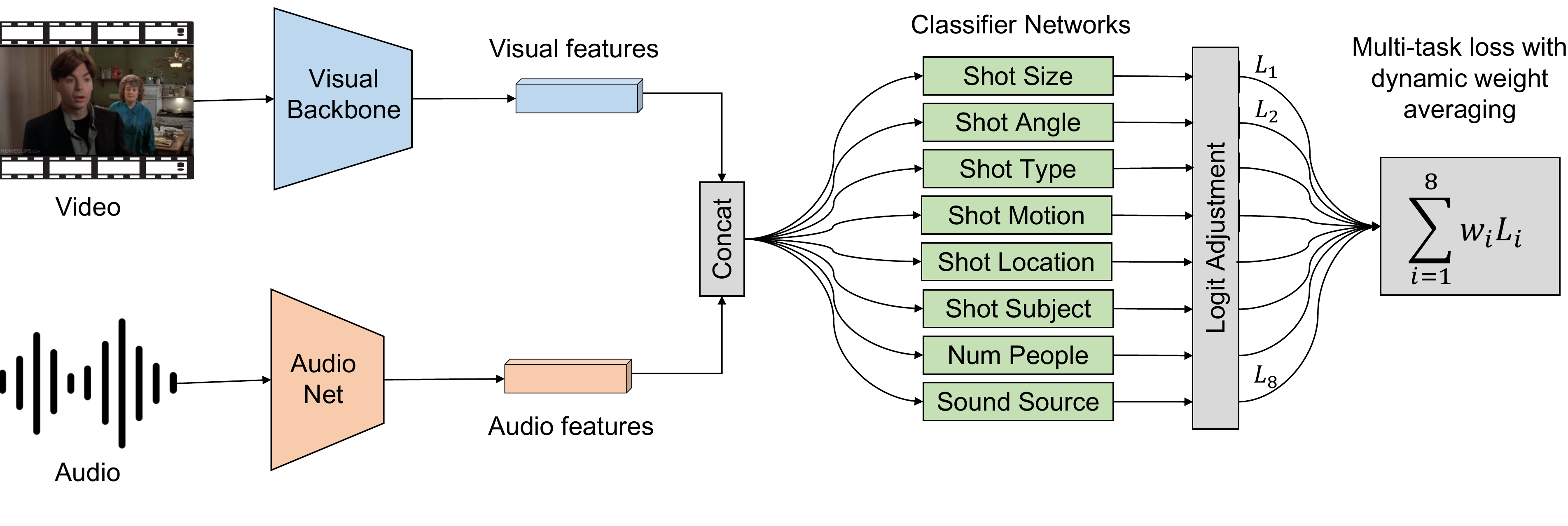}
     \vspace{-5.3mm}
    \caption{\small Overview of shot attributes classification framework. Given a shot clip, we first extract audio-visual features using a common backbone network. The extracted features are then feed into eight classifiers to predict the attributes of the shot.}
    \label{fig:task1}
    \vspace{-3mm}
\end{figure*}
\vspace{-4mm}
\subsection{Shot Attributes Classification}
\vspace{-2mm}
\label{sec:sap}
Given a shot $\calS$, shot attributes classification aims to predict the attributes of $\calS$: \texttt{shot size}, \texttt{shot angle}, \texttt{shot type}, \texttt{shot motion}, \texttt{shot location}, \texttt{shot subject},\texttt{ number of people} and \texttt{sound source} as discussed in \Sref{sec:shot_attributes}. This task would be useful for an editor to automatically classify and organize shots by their attributes during the editing process. It can also be coupled with other tasks such as video content understanding \cite{huang2020movienet,bain2020condensed,wu2021towards,liu2021multi} to establish a user query system, where shots can be retrieved by their content and attributes.

As each attribute has its own classes independent of others, shot attributes classification can be defined in a \textit{multi-task} setting, where multiple classifiers are jointly optimized together in an end-to-end manner. We design a general framework for shot attributes classification by cascading an audio-visual encoder network with eight classifier networks as shown in \Fref{fig:task1}. We use \texttt{ResNet-101}~\cite{he2016deep} and \texttt{R-3D}~\cite{tran2018closer} as visual backbone networks to extract features from image and video inputs, respectively. To incorporate features from the audio input, we design a network called \texttt{AudioNet}, which is a feed-forward network with 3 convolutional and 2 linear layers. The features extracted from different input representations are then cascaded to obtain a cross-modal feature via concatenation. This cross-modal feature is then fed into each classifier network which outputs the predicted class for the respective shot attribute. Each classifier is a simple network with 2 linear layers and ReLU activation units.

Shot attributes inherently exhibit a long tail label distribution. For instance, \textit{medium} is the most common \texttt{shot size} in movie scenes while \textit{extreme close-up} is hardly used. This imbalance makes naive training to be biased toward the dominant label~\cite{menon2020long,kang2019decoupling,tan2020equalization}. To address this problem, we implement the idea of logit adjustment~\cite{menon2020long} on each classifier output according to the label frequencies in the respective shot attribute. The training loss for our network is defined as an aggregate of the cross entropy losses from the different classifiers. To better explore the cross-task correlation during training, we use dynamic weight averaging technique~\cite{liu2019end} to scale the loss of each task (attribute) during training. 
\vspace{-3mm}

\subsection{Camera Setup Clustering}
\label{sec:sc}
\vspace{-1mm}
 Movie scenes in general contain highly frequent shot cuts as they are professionally edited by connecting several shots captured using a multi-camera system. This phenomenon can also be observed in the proposed dataset (see \Tref{tbl:data_stat}) which contains approximately 35 shots and 6 camera setups per scene. Given a list of shots $\{\calS_1, \calS_2, \ldots, \calS_k\}$ in any order, shot clustering is defined as grouping shots that belong to the same camera setup. This task could be useful during the editing process in order to catalog different shots of a scene or various takes of a particular shot into the respective camera setup they belong to.
 
We formulate this task as a high-dimensional feature clustering problem. We extract features from the given shot set and evaluate the performance of several state-of-the-art clustering algorithms. We use both traditional, \ie~\texttt{SIFT}~\cite{lowe2004distinctive}, and learning-based, \ie~ \texttt{ResNet-101}~\cite{he2016deep}, \texttt{CLIP}~\cite{radford2021learning} and \texttt{R-3D}~\cite{tran2018closer}, feature extraction methods. To establish baselines for shot clustering task, we experiment with standard clustering algorithms such as \texttt{K-Means}~\cite{lloyd1982least}, Hierarchical Agglomerative Clustering (\texttt{HAC})~\cite{mullner2011modern}, \texttt{OPTICS}~\cite{ankerst1999optics}, but also novel methods such as \texttt{FINCH}~\cite{sarfraz2019efficient}. 
 \begin{figure*}[!t]
    \centering
    \includegraphics[width=1.0\linewidth,trim={4.1cm 14cm 18.4cm 7.8cm},clip]{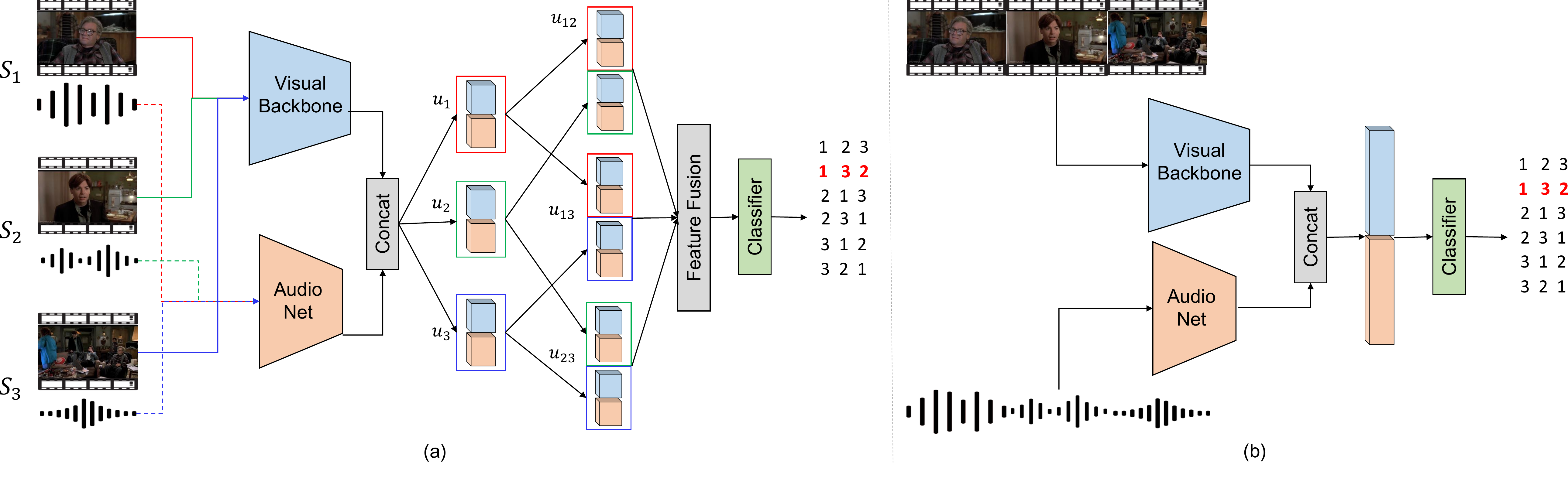}
    \vspace{-5mm}
    \caption{\small Shot ordering baselines. (a) Late feature fusion, where we first extract features from each shot in the sequence and then combine the features at later stage (b) Early input fusion, where the input shot clips are first concatenated before extracting feature.}
    \label{fig:task3}
    \vspace{-3mm}
\end{figure*}
\vspace{-3mm}
\subsection{Shot Sequence Ordering}
\label{sec:so}
\vspace{-1mm}
Shots are fundamental units in the filmmaking process. Film editors create scenes by assembling shots in a coherent pattern that best depicts the story of the scene. As previous studies~\cite{baxter2013comparing,wu2018thinking,wu2016analysing,leake2017computational} have indicated, several factors go into the selection and sequencing of shots, which can be highly subjective at times. In this task, we aim to learn general shot ordering patterns in movie scenes following a data-driven approach, where we break down a movie scene into shots, randomly shuffle the shots, and target to reconstruct the movie scene by reordering the shuffled shots. Given a sequence of \textit{contiguous} but randomly shuffled shots, \ie~$\texttt{rand}\{\calS_1, \calS_2, \ldots, \calS_k\}$, shot ordering can be formulated as a classification task. If a given scene has $k$ shots, there are $k!$ (factorial of $k$) possible ways of ordering them, \ie~$k!$ classes. We set $k = 3$ at a time for convenience and define shot order prediction as a 6-way classification problem. 

We experiment with two types of baselines for shot order prediction as shown in \Fref{fig:task3}. First, we follow previous video representation learning works~\cite{xu2011using,9204376} and perform late feature fusion, where we extract features from the input shot clips and then perform a hierarchical fusion of features. The combined features are then passed to a classifier network to predict the order (see \Fref{fig:task3}a). As a second baseline, we perform early fusion at an input level where we concatenated the shot clips and then extract features from the resulting input (see \Fref{fig:task3}b).

Compared to existing works on video clips order prediction~\cite{xu2011using,9204376,el2019skip,kim2019self}, shot ordering is a more challenging problem for two main reasons. First, previous works mostly deal with ordering different segments of a video with a single camera setup. This makes it convenient to exploit semantic and geometric correspondences across clips to analyze their temporal coherence. In contrast, the neighboring clips in a given shot sequence are often from different camera setups and there is much less content overlap across inputs making it very challenging to learn ordering patterns from only the visual stream. Second, in previous works~\cite{xu2011using,9204376,el2019skip,kim2019self} problem formulation, there is always a unique solution for ordering the input clips given that the interval between the clips is not too large. In comparison, due to the subjective (and artistic) nature of the task, there could be multiple ways of ordering shot clips in a movie scene ~\cite{baxter2013comparing,wu2018thinking,wu2016analysing,leake2017computational}.
 \begin{figure*}[!t]
    \centering
    \includegraphics[width=1.0\linewidth,trim={2.2cm 4.5cm 14.7cm 12.7cm},clip]{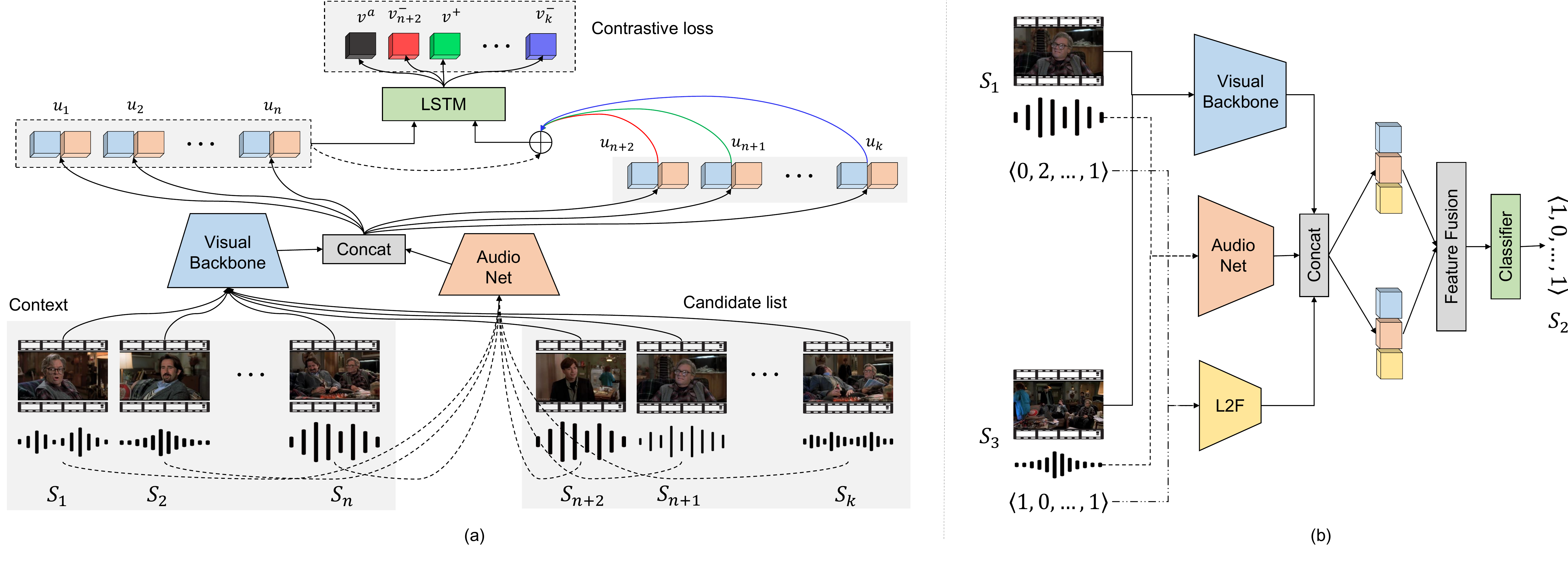}
    \caption{\small (a) Overview of next shot selection. We use a contrastive approach to learn positive (matching) and negative (non-matching) shot sequence patterns. (b) Missing shot attributes prediction. We predict the attributes of an intermediate shot given its left and right neighboring shots.}
    \vspace{-3mm}
    \label{fig:task4}
\end{figure*}
\vspace{-3mm}
\subsection{Next Shot Selection}
\label{sec:nss}
\vspace{-1mm}
During film editing, the process of assembling shots occurs in a sequential manner. Given a partial sequence of shots as a context, this task aims to anticipate the next shot from the list of available shots. Let $\{\calS_1, \calS_2, \ldots, \calS_n\}$ denote the sequence of $n$ shots provided as a history and $\texttt{rand}\{\calS_{n+1}, \calS_{n+2}. \ldots, \calS_k\}$ represent the list of $k-n$ possible candidates to follow $\calS_n$. Next shot selection is then defined as a multiple choice problem where a model makes the decision based on the affinity of each candidate shot to the previous sequence.

We formulate this task following a contrastive learning approach. First, we extract features from each shot in the given context and candidate list using an \texttt{audio-visual encoder} network (see \Fref{fig:task4}). It is important to learn the shot sequence pattern from the given context in order to anticipate the next shot candidate. Thus, we feed the extracted feature sequence $\{u_1,u_2, ..., u_n\}$ into an \texttt{LSTM}~\cite{hochreiter1997long} module, 
\ie~ $v^a = \texttt{LSTM}\{u_1, u_2, \ldots, u_n\}$. The output of the \texttt{LSTM} module is defined as an \textit{anchor} feature $v^a$ which represents an embedding for the context sequence. To generate a positive (matching) feature $v^+$  during training, we cascade the context and the correct next shot $\calS_{n+1}$ and input the resulting sequence into the \texttt{LSTM} module, \ie~$v^+ = \texttt{LSTM}\{u_1, u_2, \ldots, u_n, u_{n+1}\}$. The intuition here is to learn that $\{\calS_1, \calS_2, \ldots, \calS_n\}$ and $\{\calS_1, \calS_2, \ldots, \calS_n, \calS_{n+1}\}$ are feasible shot sequence patterns that appear in actual movie scenes. In contrast, the negative (non-matching) features are created by cascading any shot \textbf{except} $\calS_{n+1}$ to the context, \ie~$\{v^-_i\}_{i=1}^{} = \texttt{LSTM}\{u_1, u_2, \ldots, u_n, u_{i}\}$, where $i\ne n+1$. We experiment with two types of negative samples generation: i) \texttt{in-sequence}, where $k-n-1$ negatives are sampled per each input using all incorrect choices in the respective candidate list and ii) \texttt{in-batch}, where all other shot sequences in the batch are additionally used, \ie~$bk-n-1$ negative samples per each input for a batch size of $b$. We use the supervised NT-Xent loss~\cite{chen2020simple,NEURIPS2020_d89a66c7} to train our network. During inference, we compute the affinity score, the cosine similarity, between the anchor feature $v^a$ and  the feature $v^c = \texttt{LSTM}\{u_1, u_2, \ldots, u_n, u_{c}\}$, where $c$ is an index for a shot in the candidate list. The shot with the highest score is then selected as the next shot.
\vspace{-3mm}
\subsection{Missing Shot Attributes Prediction}
\label{sec:msap}
\vspace{-1mm}
With the goal of learning editing patterns in movie scenes, here, we define another task that aims to predict the attributes of a missing shot in a given incomplete shot sequence. This is different from the task in \Sref{sec:nss} as it targets to predict the likely attributes of the shot that best completes the given sequence irrespective of the shot availability. Let $\{\calS_1, \calS_2, \ldots, \calS_k\}$ denote a given incomplete sequence without $\calS_i$, where $1 < i < k$. Then, missing shot attribute prediction is formulated as a classification problem, where we predict the attributes of $\calS_i$ using the input sequence of shots and their attributes as a context. In this work, we consider a simple setup with $k = 3$, \ie~$\{\calS_1,\calS_3\}$ is given as an input and we predict the attributes of $\calS_2$. A more generalized formulation with longer sequences and missing shot(s) at arbitrary time steps is left for future work.

\Fref{fig:task4}b depicts the framework for missing shot attribute prediction. We first extract features from $\calS_1$ and $\calS_3$ using a pretrained backbone network. The attributes of $\calS_1$ and $\calS_2$ are also added as an input. For this purpose, we designed a simple 1-layered linear network called \texttt{label-2-feature (L2F)} which transforms an attribute vector of size 8 into a feature embedding. The extracted features are then concatenated and fed into multiple classifier networks which predict the different attributes of the missing shot $\calS_2$. Like the task in \Sref{sec:sap}, logit adjustment is applied to the output of the classifiers  during training to prevent the network from being biased to the dominant labels. The training is done in a multi-task setting using dynamic weight averaging~\cite{liu2019end} to scale the cross entropy losses from the different classifiers. 

\vspace{-3mm}
\section{Experimental Results and Discussion}
\label{sec:experiment}

\subsubsection{Dataset.}
We follow a $\texttt{train-val-test}$ scene split of 70-10-20 in all experiments (see \Tref{tbl:data_stat}). As the scenes in the proposed dataset are non-overlapping, the train, validation, and test splits are disjoint sets. For the shot attributes classification task, we use all the shots in the respective scene split for training and evaluation. For shot ordering and missing shot attributes prediction tasks, we generate train, validation, and test sets by sampling 3 consecutive shots from a scene at a time. For the next shot selection task, on the other hand, we sample 9 consecutive shots at a time. The first 4 shots in the sequence are used as a context. The remaining 5 shots are used to make a candidate list. 
\vspace{-5mm}
\subsubsection{Evaluation metrics.} For shot attributes classification and missing shot attributes prediction tasks, we report the \texttt{average per-class accuracy} to take the long tail distribution problem into account. For shot ordering and next shot selection tasks, we simply evaluate the \texttt{overall accuracy}. For shot clustering task, we evaluate the quality of the generated clusters with respect to ground truth clusters on 3 different metrics: Purity score (\texttt{PS}), Normalized mutual information (\texttt{NMI}) and Rand index (\texttt{RI})~\cite{schutze2008introduction}. 
\vspace{-5mm}
\subsubsection{Implementation Details.} We use \texttt{ResNet-101}~\cite{he2016deep} and \texttt{R-3D}~\cite{tran2018closer} as visual backbone networks to extract features from image and video inputs, respectively. In all experiments, the backbone network is initialized with pretrained weights (\texttt{ResNet-101} - pretrained on ImageNet~\cite{deng2009imagenet} and \texttt{R-3D} - pretrained on Kinetics-400~\cite{kay2017kinetics}) and fine-tuned during training. We uniformly sample 16 frames from a shot clip as an input to \texttt{R-3D}, while we use the central frame of a shot for \texttt{ResNet-101}. Refer to the supplementary for task-level implementation details.

\begin{table*}[!t]
\begin{center}
\caption{\small Quantitative analysis on shot attributes classification.}
\vspace{-2.5mm}
\label{tbl:shot_att_audvid}
\setlength{\tabcolsep}{4.5pt}
\renewcommand{\arraystretch}{0.85}

\mytabular{
\begin{tabular}{l|cccc|cccc|cccc}
\toprule
& \multicolumn{8}{c|}{\textbf{Multi-task training}} & \multicolumn{4}{c}{\textbf{Single-task training}}\\ \cmidrule(lr){2-9} \cmidrule(lr){10-13}
& \multicolumn{4}{c|}{\texttt{Video}} & \multicolumn{4}{c|}{\texttt{Video + Audio}} & \multicolumn{4}{c}{\texttt{Video + Audio}} \\ \cmidrule(lr){2-5}\cmidrule(lr){6-9}\cmidrule(lr){10-13}
&\multicolumn{2}{c}{Nai\"ve} & \multicolumn{2}{c}{Logit adj.} & \multicolumn{2}{c}{Nai\"ve} & \multicolumn{2}{c}{Logit adj.} & \multicolumn{2}{c}{Nai\"ve} & \multicolumn{2}{c}{Logit adj.}\\ \cmidrule(lr){2-3} \cmidrule(lr){4-5} \cmidrule(lr){6-7} \cmidrule(lr){8-9} \cmidrule(lr){10-11} \cmidrule(lr){12-13}
Attribute &\texttt{Val} & \texttt{Test} & \texttt{Val} & \texttt{Test} & \texttt{Val} & \texttt{Test} & \texttt{Val} & \texttt{Test} & \texttt{Val} & \texttt{Test} & \texttt{Val} & \texttt{Test} \\ \midrule

\texttt{Shot size} & 35.7 & 35.5 & 67.8 & 66.9 & 36.2 & 36.4 & 66.8 & 65.0 & 38.7 & 39.1 & 70.9 & 67.6\\
\texttt{Shot angle} & 25.8 & 25.8 & 62.2 & 53.2 & 27.6 & 27.7 & 58.6 & 49.5 & 29.1 & 28.9 & 63.0 & 49.8\\
\texttt{Shot type} & 59.5 & 60.8 & 63.9 & 64.9 & 59.8 & 61.0 & 63.7 & 65.3 & 60.1 & 62.3 & 64.9 & 66.7\\
\texttt{Shot motion} &32.1 & 31.7 & 42.8 & 42.7 &32.3 & 31.8 & 44.6 & 43.2 & 32.3 & 31.2 & 47.4 & 43.7 \\
\texttt{Shot location} & 82.9 & 81.9 & 84.4 & 83.3 &83.0 & 80.9 & 83.7 & 83.7 & 83.4 & 82.8 & 83.9 &  84.0\\
\texttt{Shot subject} & 40.0 & 39.8 & 50.8 & 47.4 &40.0 & 39.7 & 50.2 & 46.7 & 42.2 & 40.7 & 54.2 & 48.0 \\
\texttt{Num. of people} & 55.0 & 55.1 & 61.3 & 61.2&55.1 & 55.3 & 60.9 & 61.4& 56.1 & 57.1 & 61.5 & 63.5\\
\texttt{Sound source} & 25.0 & 25.0 & 34.4 & 32.6 & 25.0 & 25.0 & 41.0 & 38.9 & 25.0 & 25.0 & 38.1 & 35.3\\ \midrule
\texttt{Average} & 44.5 & 44.4 & \textbf{58.4} & \textbf{56.5} & 44.9 & 44.7 & \textbf{58.7} & \textbf{56.7} & 45.9 & 45.9 & \textbf{60.5} & \textbf{57.3} \\
\bottomrule
\end{tabular}
}
\end{center}
\vspace{-6mm}
\end{table*}

\vspace{-3mm}
\subsection{Experimental Results}
\subsubsection{Shot Attributes Classification.} In \Tref{tbl:shot_att_audvid}, we present the performance of our network trained in two different settings: i. \texttt{multi-task}, where all eight classifiers are jointly trained together in an end-to-end manner, and ii. \texttt{single-task}, where one classifier is optimized at a time. As can be inferred from \Tref{tbl:shot_att_audvid}, individually training each classifier generally results in a better accuracy compared to training all classifiers together. It can also be noticed from \Tref{tbl:shot_att_audvid} that taking the long tail distribution problem into account during training consistently leads to a significantly better result in comparison with nai\"ve training. For example, in a single-task setting, applying logit adjustment ({\small Logit adj.}) during training results in a $31.8\%$ and $24.8\%$ average accuracy improvement on validation and test sets, respectively. For attributes with imbalanced label distributions such as \texttt{shot size} and \texttt{shot angle}, we have observed that nai\"vely trained network performs very well for the dominant classes but extremely poorly for low-frequency classes. On the other hand, a network trained with logit adjustment gives a relatively balanced per-class accuracy, and hence better overall performance. 

We use different types of input representations for shot attributes classification. \Tref{tbl:shot_att_audvid} compares their performance. Using video features as an input gives a competitive performance for most shot attributes except for \texttt{sound source}, where adding audio features results in a $19.3\%$ performance boost on both validation and test sets. Our baselines achieve a relatively low accuracy when predicting \texttt{shot motion} and \texttt{sound source}. The lack of explicit modeling of motion could be one contributing factor to the low performance on \texttt{shot motion}~\cite{rao2020unified}. The low accuracy \texttt{sound source} classification is most likely due to the fine-grained nature of the task. For instance, it could be very ambiguous to differentiate between \texttt{on-screen}, \texttt{off-screen} and \texttt{external-narration} classes. Incorporating motion information in the form of optical flow along with other input modalities and exploring the correlation of audio and visual features with a carefully designed attention mechanism are interesting research directions.

\begin{table*}[!t]
\begin{center}
\caption{Quantitative analysis on camera setup clustering.}
\vspace{-2.5mm}
\label{tbl:shot_cluster}
\setlength{\tabcolsep}{3.5pt}
\renewcommand{\arraystretch}{0.85}

\mytabular{
\begin{tabular}{l|ccc|ccc|ccc|ccc}
\toprule
& \multicolumn{6}{c|}{\textbf{Parameter-based}} & \multicolumn{6}{c}{\textbf{Parameter-free}}\\\cmidrule(lr){2-7} \cmidrule(lr){8-13}
& \multicolumn{3}{c|}{K-Means~\cite{lloyd1982least}} & \multicolumn{3}{c|}{HAC~\cite{mullner2011modern}} & \multicolumn{3}{c|}{OPTICS~\cite{ankerst1999optics}} & \multicolumn{3}{c}{FINCH~\cite{sarfraz2019efficient}}\\ 
\cmidrule(lr){2-4} \cmidrule{5-7} \cmidrule(lr){8-10} \cmidrule(lr){11-13}
Method & \texttt{RI} & \texttt{NMI} & \texttt{PS} & \texttt{RI} & \texttt{NMI} & \texttt{PS} & \texttt{RI} & \texttt{NMI} & \texttt{PS} & \texttt{RI} & \texttt{NMI} & \texttt{PS} \\ \midrule
\texttt{SIFT}~\cite{lowe2004distinctive} & 0.863 & 0.779 & 0.837 & 0.865 & 0.785 & 0.840 & 0.786 & 0.685 & 0.790 & 0.795 & 0.692 & 0.777\\ \midrule
\texttt{ResNet-101}~\cite{he2016deep}  & \textbf{0.947}
& \textbf{0.913} & \textbf{0.937} & \textbf{0.946} & \textbf{0.914} & \textbf{0.939} & 0.784 & 0.718 & 0.801 & 0.887 & 0.834
& 0.887 \\
\texttt{CLIP}~\cite{radford2021learning}  & 0.907 & 0.858 & 0.894 & 0.912 & 0.867 & 0.899 & 0.836 & 0.769 & 0.865 & 0.845 & 0.780 & 0.851\\
\texttt{\texttt{ResNet-101} (Ours)} & 0.921 & 0.873 & 0.908 & 0.922 & 0.876 & 0.909 & \textbf{0.868} & \textbf{0.802} & \textbf{0.889} & \textbf{0.889} & \textbf{0.838} & \textbf{0.895}\\ \midrule
\texttt{R-3D}~\cite{tran2018closer} & 0.906 & 0.846 & 0.891 & 0.910 & 0.856 & 0.897 & 0.693 & 0.601 & 0.723 & 0.814 & 0.728 & 0.813 \\
\texttt{R-3D (Ours)} & 0.902 & 0.840 & 0.882 & 0.903 & 0.844 & 0.884 & 0.850 & 0.775 & 0.871 & 0.872 & 0.805 & 0.866\\ 
 \bottomrule
\end{tabular}
}
\end{center}
\vspace{-5mm}
\end{table*}

\vspace{-3mm}
\subsubsection{Camera Setup Clustering.} We perform scene-level shot clustering, where we group the shots of a given scene into different camera setups. The averaged results for all scenes in the dataset are summarized in \Tref{tbl:shot_cluster}. We use image-based and video-based feature extraction methods to establish baselines. We also compare the visual backbone of our framework trained on shot attribute prediction task in \Sref{sec:shot_attributes}. For clustering, we experiment with four standard clustering algorithms. \texttt{K-Means}~\cite{lloyd1982least} and \texttt{HAC}~\cite{mullner2011modern} require the number of clusters as an input parameter, \ie~\textit{parameter-based}, while \texttt{OPTICS}~\cite{ankerst1999optics} and \texttt{FINCH}~\cite{sarfraz2019efficient} generate clusters without relying on the number of clusters as an input, \ie~\textit{parameter-free}. 

\Tref{tbl:shot_cluster} shows that image-based feature extraction methods generally perform better than video-based backbones. It is also worth noting that, for parameter-based clustering, \texttt{ResNet-101}~\cite{he2016deep} pretrained on  ImageNet~\cite{deng2009imagenet} achieves the highest clustering accuracy on all metrics. However, for parameter-free clustering, \texttt{ResNet-101} pretrained on the proposed dataset consistently outperforms \texttt{ResNet-101} (pretrained on  ImageNet). The same notion can also be observed for \texttt{R-3D}~\cite{tran2018closer} which is pretrained on Kinetics-400~\cite{kay2017kinetics}. These results highlight that shot attributes classification could be used as a pretext task for better shot clustering as the ground truth for the number of clusters is not generally available.

\begin{table*}[!t]
\begin{center}
\caption{\small Quantitative analysis on shot sequence ordering.}
\vspace{-2.5mm}
\label{tbl:shot_ord}
\setlength{\tabcolsep}{4.5pt}
\renewcommand{\arraystretch}{0.85}

\mytabular{
\begin{tabular}{l|ccc|ccc|ccc}
\toprule
& \multicolumn{3}{c|}{\textbf{Frame}} & \multicolumn{3}{c|}{\textbf{Video}} & \multicolumn{3}{c}{\textbf{Audio + Video}}\\ \cmidrule(lr){2-4} \cmidrule(lr){5-7} \cmidrule(lr){8-10}
Method & \texttt{Val} & \texttt{Test} & \texttt{Survey} & \texttt{Val} & \texttt{Test} & \texttt{Survey} & \texttt{Val} & \texttt{Test} & \texttt{Survey} \\ \midrule
\texttt{Random} & 16.6 & 16.6 & 16.6 & 16.6 & 16.6 & 16.6 & 16.6 & 16.6 & 16.6\\\midrule
\texttt{Baseline-I} & 21.0 & 21.5 & 20.0 & 21.7 & 22.4 & 23.3 & 23.1 & 24.4 & 26.7 \\
\texttt{Baseline-II } & 25.0 & 25.7 & 26.7 & 26.5 & 27.4 & 33.3 & \textbf{29.3} & \textbf{30.7} & 33.3 \\ \midrule
\texttt{Human} & - & - & 32.0 & - & - & 39.9 & - & - & \textbf{55.6}\\
\toprule
\textbf{Cinematography patterns} & & & & & & & & & \\ \cmidrule(lr){1-1}
\texttt{Baseline-I (insert)} & 26.2 & 25.8 & - & 27.9 & 27.5 & - & 30.1 & 29.8 & - \\
\texttt{Baseline-II (insert)} & 39.6 & 37.3 & - & 41.5 & 38.4 & - & 42.2 & 39.5 & - \\ \midrule
\texttt{Baseline-I (intensify)} & 29.6 & 30.1 & - & 30.7 & 31.4 & - & 32.2 & 33.1 & -\\
\texttt{Baseline-II (intensify)} & 34.1 & 35.8 & - & 38.1 & 40.5 & - & 44.2 & 48.0 & - \\ 
\bottomrule
\end{tabular}
}
\end{center}
\vspace{-3mm}
\end{table*}

\vspace{-3mm}
\subsubsection{Shot Sequence Ordering.} The results for the shot order prediction task are presented in \Tref{tbl:shot_ord}. We evaluate two baselines for shot order prediction: i. \texttt{Baseline-I}, where we first extract features from each shot in the sequence and then perform hierarchical feature fusion in the later stage as shown in \Fref{fig:task3}a, and ii. \texttt{Baseline-II}, where we first concatenate the shots in the given input and then extract features from the resulting sequence (see \Fref{fig:task3}b). We evaluate the two baselines for different input representations. As shot ordering is a 6-way classification problem, random order prediction has an accuracy of $16.6\%$. As can be seen from \Tref{tbl:shot_ord}, the best performance for both baselines is achieved when using \texttt{audio + video} compared to using only \texttt{video} or a single \texttt{frame} as an input. This is intuitive as audio-visual features provide a richer context for the network to find correspondence between the shots when predicting order. 

\Tref{tbl:shot_ord} also shows that early fusion of inputs leads to significantly better results compared to late feature fusion. For example, when using \texttt{audio + video}, \texttt{Baseline-II} outperforms \texttt{Baseline-I} by a margin of 26.8\% and 25.8\% on validation and test sets, respectively. This is mainly because early fusion enables the model to implicitly learn the correlation between shots at different levels of abstraction since all inputs are processed simultaneously. Instead, late feature fusion only learns limited correspondences as each shot is encoded independently.

The results in \Tref{tbl:shot_ord}  are low mainly because predicting the order of shots is a very challenging problem. As discussed in \Sref{sec:so}, the shots in the input sequence are often from different camera setups, and hence, it is very difficult to exploit semantic and geometric correspondences which are crucial to learning order. To further analyze the performance of our models in comparison with humans, we conducted a survey. We sampled 30 triplets from the test set and asked more than 160 people to predict the order of the randomly shuffled shots, where each shot is represented using a single \texttt{frame}, \texttt{video} and \texttt{video + audio}. To prevent people from exploiting very noticeable transitions between shots, we embedded blank frames between the shot clips. 

As can be concluded from \Tref{tbl:shot_ord}, despite the better accuracy compared to our baselines, 
the task of ordering shots is also difficult for humans particularly when the shots are represented using a \texttt{frame} or \texttt{video}. The significant performance surge for humans in \texttt{audio + video} setting is most likely because humans can comprehend the content of the audio. For instance, if there is a dialogue between subjects in the given shots, humans can easily establish the order of the shots based on the speech of the subjects. An exciting future work would be to further exploit the content of the audio in the form of speech or other relevant representations in order to imitate human comprehension.

Although a broadly formulated shot ordering problem could be challenging, we analyze the performance of our baselines for shot sequences that contain commonly used cinematography patterns~\cite{baxter2013comparing,wu2018thinking,wu2016analysing,leake2017computational}. First, we evaluate the \texttt{insert} pattern, where one of the shots in the sequence is an \texttt{insert} shot. In this case, the number of possibilities for ordering decreases to 2 since the \texttt{insert} shot should always be in the middle. It is worth noting that a model doesn't have this pre-existing knowledge. As can be seen from \Tref{tbl:shot_ord}, the performance of both baselines significantly increases for validation and test samples that contain an \texttt{insert} shot. The same phenomenon can also be observed for \texttt{intensify} pattern, where an editor uses a sequence of shots moving gradually closer, \ie~ decreasing \texttt{shot size}, to build up emotion \footnote{We consider 3 \texttt{intensify} patterns: \texttt{extreme-wide} - \texttt{wide} - \texttt{medium}, \texttt{wide} - \texttt{medium} - \texttt{close-up}, \texttt{medium} - \texttt{close-up} - \texttt{extreme-close-up}.}. These results highlight that our baselines have implicitly learned common cinematography patterns during training.

\begin{table}[!t]
\begin{center}
\caption{\small Quantitative analysis on next shot selection.}
\vspace{-2.5mm}
\label{tbl:next_shot}
\setlength{\tabcolsep}{4.5pt}
\renewcommand{\arraystretch}{0.85}
\mytabular{
\begin{tabular}{l|cc|cc|cc}
\toprule
& \multicolumn{2}{c|}{\textbf{Frame}} & \multicolumn{2}{c|}{\textbf{Video}} & \multicolumn{2}{c}{\textbf{Audio + Video}}\\ \cmidrule(lr){2-3} \cmidrule(lr){4-5} \cmidrule(lr){6-7}
Method & \texttt{Val} & \texttt{Test} & \texttt{Val} & \texttt{Test} & \texttt{Val} & \texttt{Test} \\ \midrule
\texttt{Random} & 20.0 & 20.0 & 20.0 & 20.0 & 20.0 & 20.0 \\
\texttt{Cosine Sim.} & 14.3 & 13.4 & 13.3 & 13.4  & - & - \\
\texttt{Ours (in-sequence)} & 34.1 & 34.0  & 37.9 & 37.5 & 39.0 & 38.7 \\
\texttt{Ours (in-batch)} & 38.4 & 38.2 & 41.3 & 41.0 & \textbf{41.6} & \textbf{41.4} \\
\bottomrule
\end{tabular}
}
\end{center}
\vspace{-5mm}
\end{table}

\vspace{-4mm}
\subsubsection{Next Shot Selection.} \Tref{tbl:next_shot} shows the results for next shot selection task. We experiment with a shot sequence length of 9 for quantitative evaluation, where the first 4 shots are used as context and the remaining 5 are randomly shuffled and fed into the network as a candidate list for the next shot. In this setup, the random chance of accurately selecting the next shot is 20\%. To further demonstrate the previously mentioned concept that the neighboring shots in a movie scene are usually from different camera setups, we use a nai\"ve \texttt{cosine similarity} between the end shot in the given context, \ie~$\calS_4$, and each shot in the candidate list as a baseline for next shot selection task. Here, we extract features from the shots using pretrained backbone networks and evaluate the cosine similarity between the extracted features. As can be seen from \Tref{tbl:next_shot}, nai\"ve \texttt{cosine similarity} performs even worse than random chance. In comparison, our proposed baselines (\Sref{sec:nss}) perform significantly better. Our approach achieves an accuracy of $41.6\%$ and $41.4\%$ on validation and test sets, respectively. We also observe that using a larger number of negatives (\texttt{in-batch}) during training improves performance. 

\vspace{-3mm}
\subsubsection{Missing Shot Attributes Prediction.} This task targets to predict the attributes of an intermediate shot given its left and right neighboring shots. In \Tref{tbl:missing_shot}, we present the results on four shot attributes, \ie~\texttt{shot size}, \texttt{shot angle}, \texttt{shot type} and \texttt{shot motion}, for a model trained in a multi-task setting. To confirm that the proposed model indeed uses the input shots as a context and does not simply converge to always predicting the dominant labels, we evaluated the accuracy of predicting the dominant label every time for each shot attribute. As can be inferred from the table, the proposed model outperforms the nai\"ve \texttt{dominant label} prediction baseline by a large margin. It can also be noticed that incorporating the attributes of the input shots along with other representations consistently improves model accuracy across all attributes.

As can be inferred from \Tref{tbl:missing_shot}, the multi-task training setup does not always lead to a balanced performance for the missing shot attributes prediction task. For instance, when using \texttt{frame} as an input, the performance gap between \texttt{shot size} and other attributes is notably large in comparison with using other input representations. This is mainly because the model overfitted to the \texttt{shot size} attribute for this particular input setup. 
\vspace{-1mm}
\begin{table*}[!t]
\begin{center}
\caption{\small Quantitative analysis on missing shot attributes prediction.}
\vspace{-2.5mm}
\label{tbl:missing_shot}
\setlength{\tabcolsep}{4.5pt}
\renewcommand{\arraystretch}{0.85}

\mytabular{
\begin{tabular}{l|cc|cc|cc|cc}
\toprule
& \multicolumn{2}{c|}{\textbf{Shot size}} & \multicolumn{2}{c|}{\textbf{Shot angle}} & \multicolumn{2}{c|}{\textbf{Shot type}} & \multicolumn{2}{c}{\textbf{Shot motion}}\\ \cmidrule(lr){2-3} \cmidrule(lr){4-5} \cmidrule(lr){6-7} \cmidrule(lr){8-9}
Method & \texttt{Val} & \texttt{Test} & \texttt{Val} & \texttt{Test} &  \texttt{Val} & \texttt{Test} & \texttt{Val} & \texttt{Test} \\ \midrule
\texttt{Dominant label} & 20.0 & 20.0 & 20.0 & 20.0 & 16.7 & 16.7 & 20.0 & 20.0\\ \midrule
\texttt{Frame} & 40.9 & 32.4 & 22.6 & 30.5 & 26.6 & 26.1 & 25.0 & 25.8 \\
\texttt{Frame + Attributes} & \textbf{47.8} & \textbf{44.6} & 28.5 & 34.1 & 32.0 & 34.5 & 31.0 & 31.8\\  \midrule
\texttt{Video} & 34.3 & 32.8 & 29.0 & 34.7 & 31.3 & 33.2 & 30.1 & 30.7\\ 

\texttt{Video + Attributes} & 38.3 & 35.3 & 30.4 & 37.2 & 32.7 & 35.0 & 31.9 & 32.1\\  \midrule
\texttt{Video + Audio} & 36.4 & 35.3 & 31.5 & 35.8 & 31.9 & 33.4 & 30.5 & 30.9 \\
\texttt{Video + Audio + Attributes} & 39.2 & 37.2 & \textbf{33.7} & \textbf{39.0} & \textbf{32.9} & \textbf{35.4} & \textbf{32.3} & \textbf{33.2}\\
\bottomrule
\end{tabular}
}
\end{center}
\vspace{-5mm}
\end{table*}
\vspace{-3mm}
\section{Conclusion}
\vspace{-2mm}
We introduced the Anatomy of Video Editing (AVE) dataset and benchmark. We gathered more than 1.5M manually labeled tags, with relevant concepts to cinematography, from 196176 shots sampled from movie scenes. We also annotated the shot transitions and camera setup in movie scenes, which allowed us to recover the scene composition. We also define five tasks to help attain research progress in automatic footage organization and assisted video assembling. We hope that our work will inspire new computer vision technologies and spur research in machine listening, speech and language understanding, and graphics. Moreover, we believe our dataset will foster the design of new relevant tasks for AI-assisted editing. For instance, our sound-source annotations can facilitate the study of music selection for video. The scene composition labels can enable tasks related to recommending pace and rhythm for cutting. 

\clearpage
%
%
\bibliographystyle{splncs04}
\bibliography{egbib}
\end{document}


\pagestyle{headings}
\mainmatter
\def\ECCVSubNumber{4736}  

\title{--- Supplementary Material --- \\The Anatomy of Video Editing: A Dataset and Benchmark Suite for AI-Assisted Video Editing} 

\titlerunning{ECCV-22 submission ID \ECCVSubNumber} 
\authorrunning{ECCV-22 submission ID \ECCVSubNumber} 
\author{Anonymous ECCV submission}
\institute{Paper ID \ECCVSubNumber}

\titlerunning{The Anatomy of Video Editing}
%
\author{Dawit Mureja Argaw\inst{1,2} \and
Fabian Caba Heilbron\inst{1} \and
Joon-Young Lee\inst{1} \and \\
Markus Woodson\inst{1} \and
In So Kweon\inst{2}}
\authorrunning{D. Mureja et al.}
%
\institute{Adobe Research \and KAIST
}
\maketitle
\begin{figure*}[!t]
    \centering
    \begin{subfigure}[]{1\textwidth}
    \centering
    \includegraphics[width = \textwidth, trim={3.2cm 0.0cm 3.20cm 0.3cm},clip]{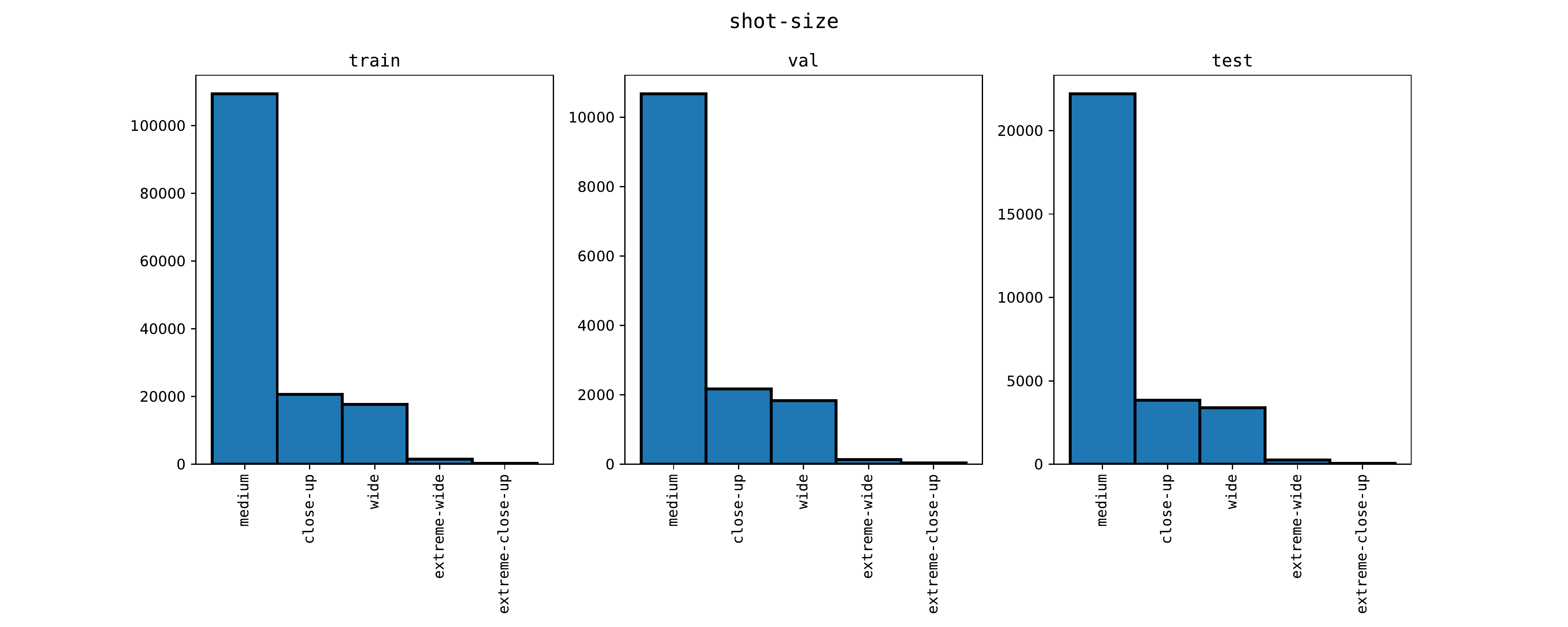}
    \end{subfigure}
    \begin{subfigure}[]{1\textwidth}
    \centering
    \includegraphics[width = \textwidth, trim={3.2cm 1.0cm 3.20cm 0.3cm},clip]{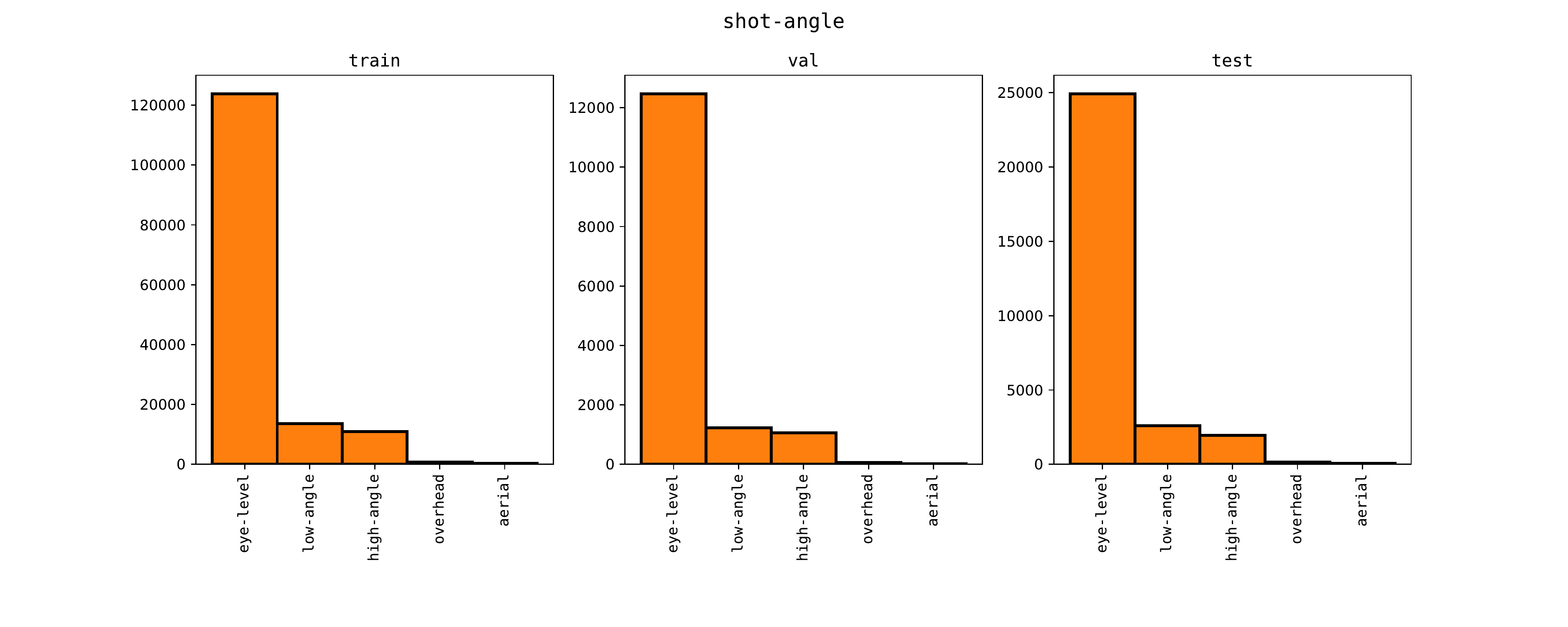}
    \end{subfigure}
    \begin{subfigure}[]{1\textwidth}
    \centering
    \includegraphics[width = \textwidth, trim={3.2cm 1.2cm 3.20cm 0.3cm},clip]{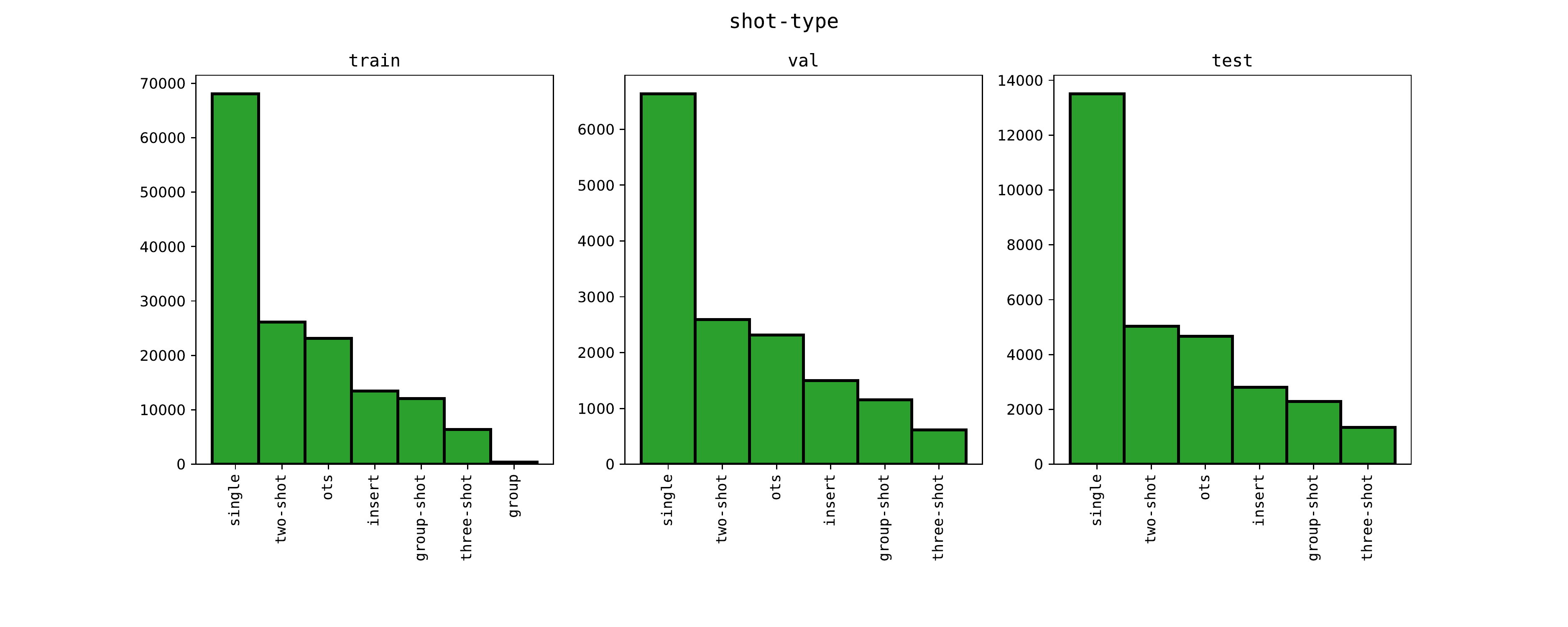}
    \end{subfigure}
    \caption{\small Class-wise distribution statistics in training, validation and testing splits for \texttt{shot size},  \texttt{shot angle} and \texttt{shot type} attributes.}
    \label{fig:qual_att1}
    \vspace{5mm}
\end{figure*}

\begin{figure*}[!t]
    \centering
    \begin{subfigure}[]{1\textwidth}
    \centering
    \includegraphics[width = \textwidth, trim={3.2cm 0.4cm 3.20cm 0.3cm},clip]{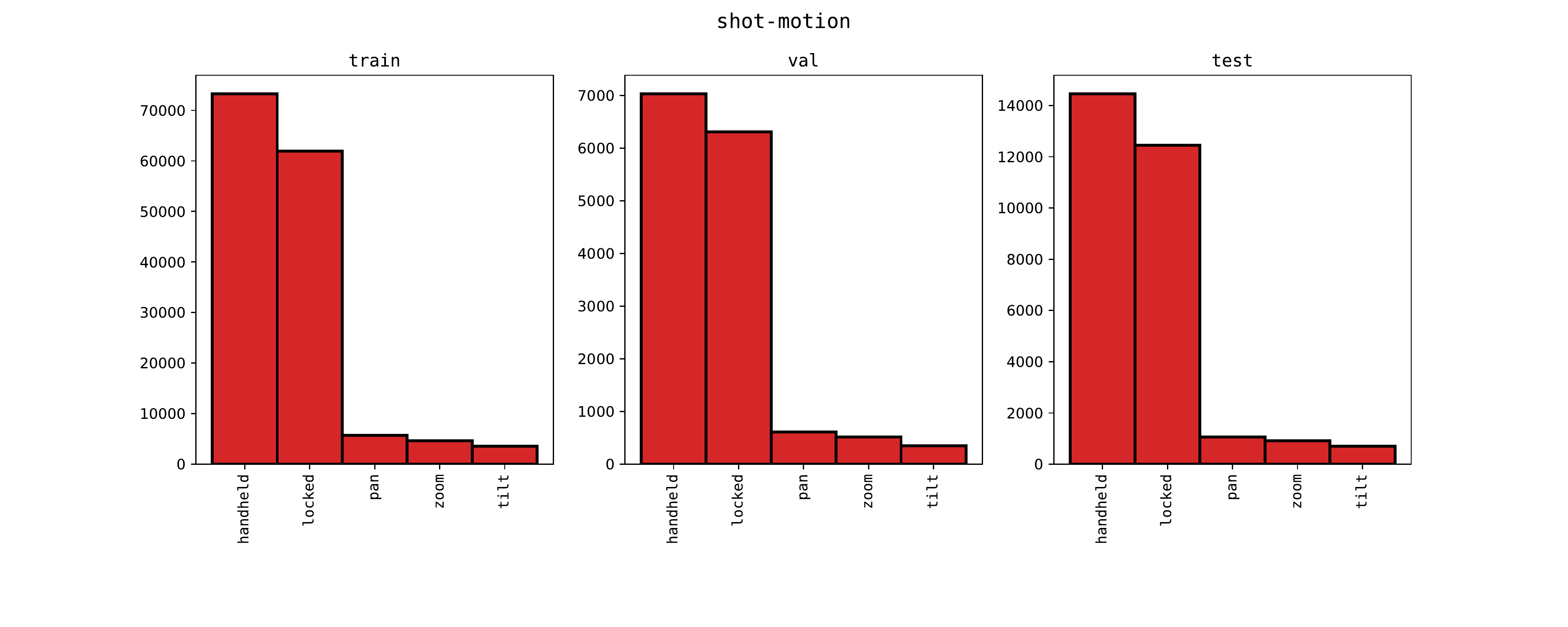}
    \end{subfigure}
    \begin{subfigure}[]{1\textwidth}
    \centering
    \includegraphics[width = \textwidth, trim={3.2cm 1.4cm 3.20cm 0.3cm},clip]{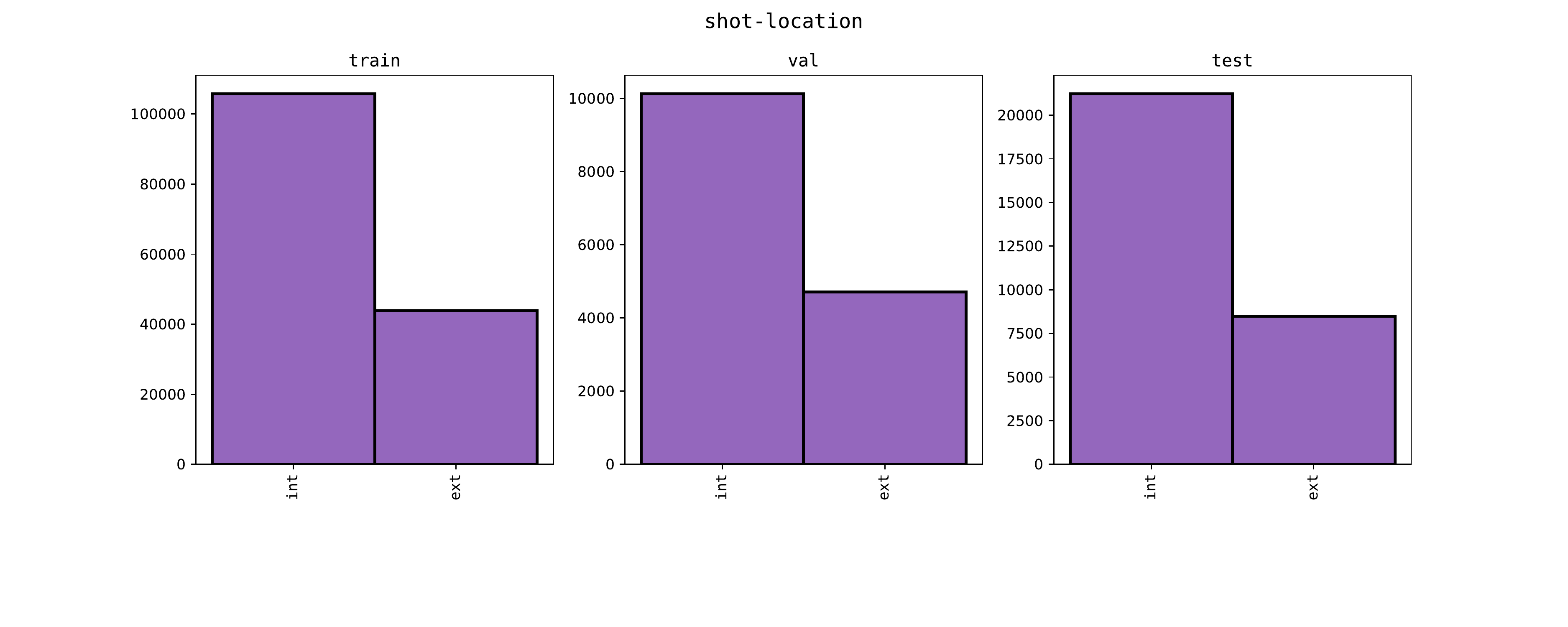}
    \end{subfigure}
    \begin{subfigure}[]{1\textwidth}
    \centering
    \includegraphics[width = \textwidth, trim={3.2cm 1.2cm 3.20cm 0.3cm},clip]{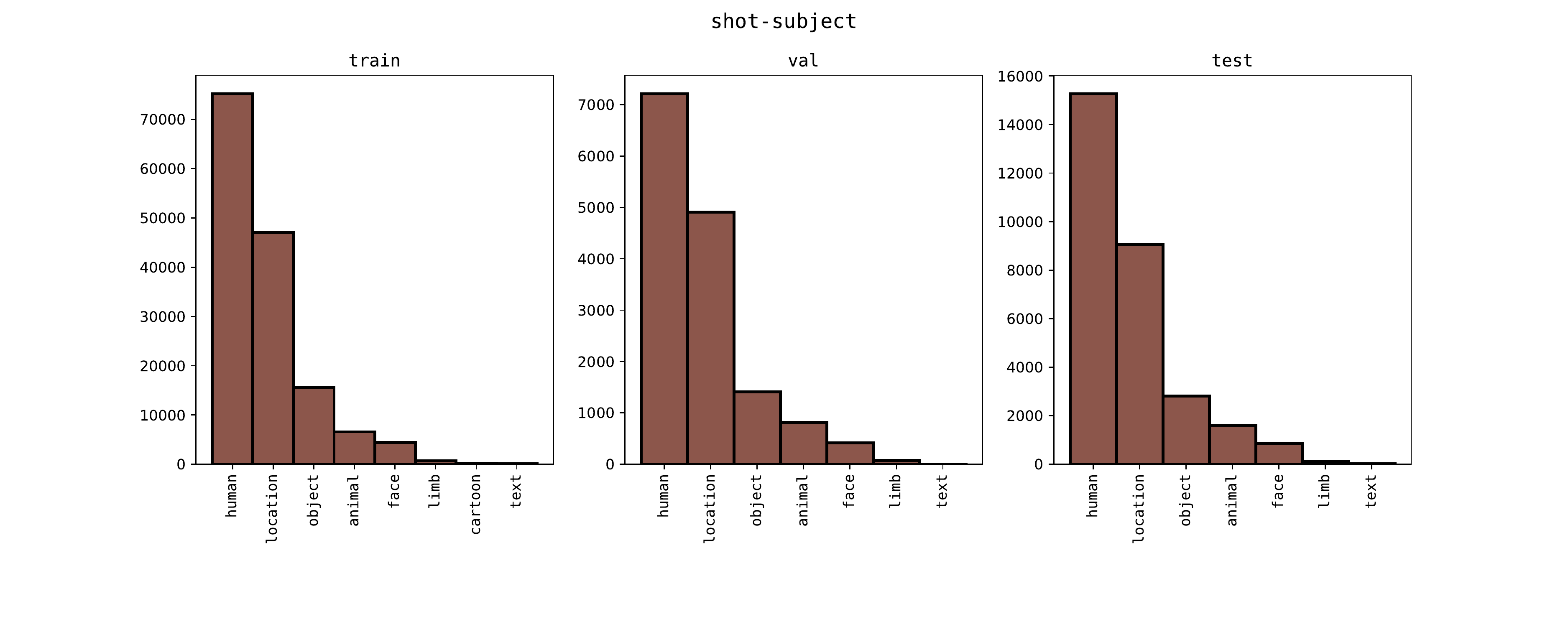}
    \end{subfigure}
    \caption{\small Class-wise distribution statistics in training, validation and testing splits for \texttt{shot motion},  \texttt{shot location} and \texttt{shot subject} attributes.}
    \label{fig:qual_att2}
    \vspace{10mm}
\end{figure*}

\begin{figure*}[!t]
    \centering
    \begin{subfigure}[]{1\textwidth}
    \centering
    \includegraphics[width = \textwidth, trim={3.2cm 2.2cm 3.20cm 0.3cm},clip]{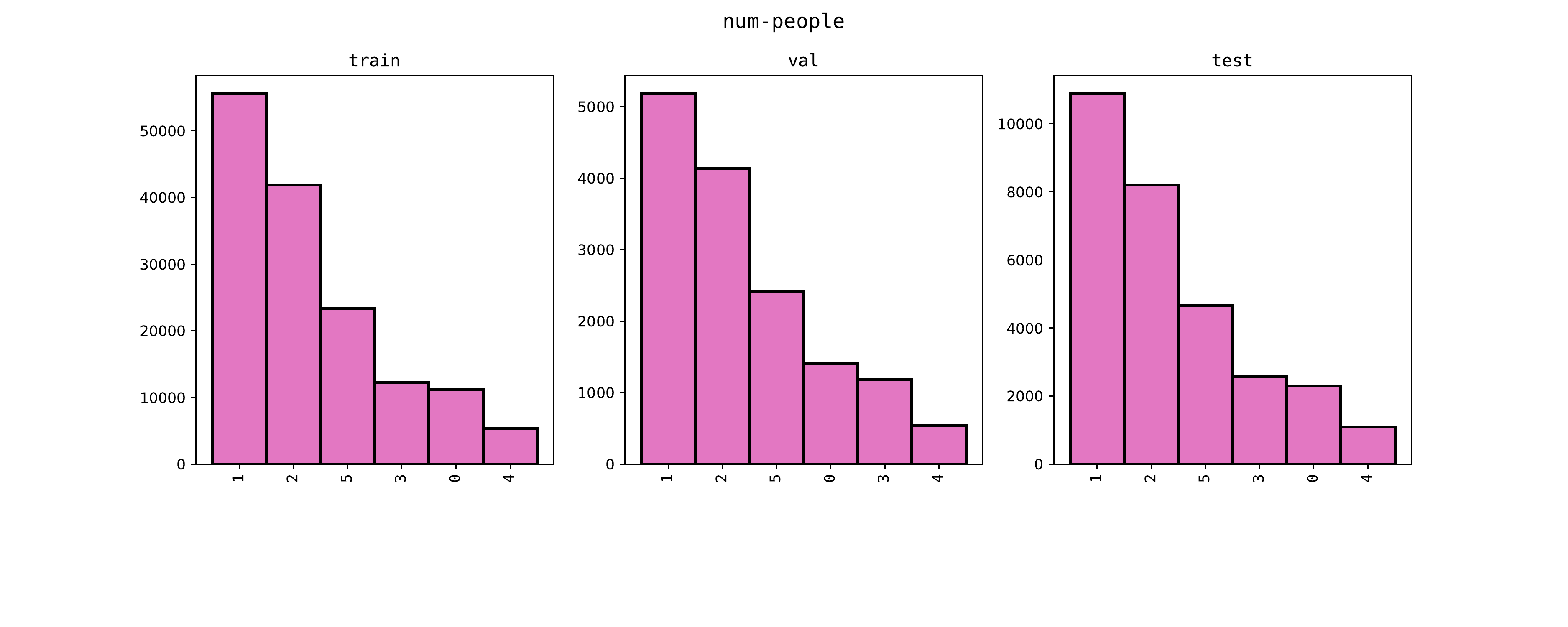}
    \end{subfigure}
    \begin{subfigure}[]{1\textwidth}
    \centering
    \includegraphics[width = \textwidth, trim={3.2cm 0.1cm 3.20cm 0.3cm},clip]{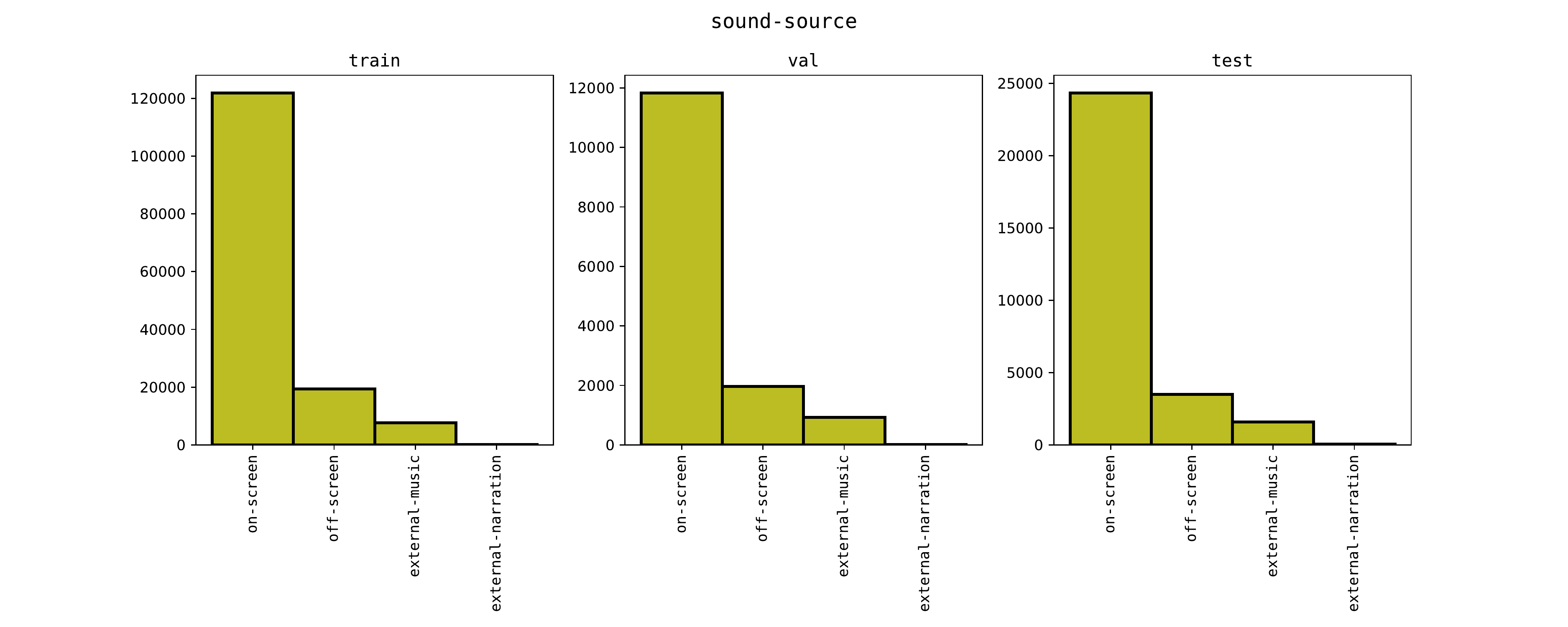}
    \end{subfigure}
    \caption{\small Class-wise distribution statistics in training, validation and testing splits for \texttt{num. of people} and \texttt{sound source} attributes.}
    \label{fig:qual_att3}
    \vspace{-2mm}
\end{figure*}

\begin{figure*}[!t]
    \centering
    \begin{subfigure}[]{0.48\textwidth}
    \centering
    \includegraphics[width = \textwidth, trim={0.8cm 0.2cm 1.60cm 0.9cm},clip]{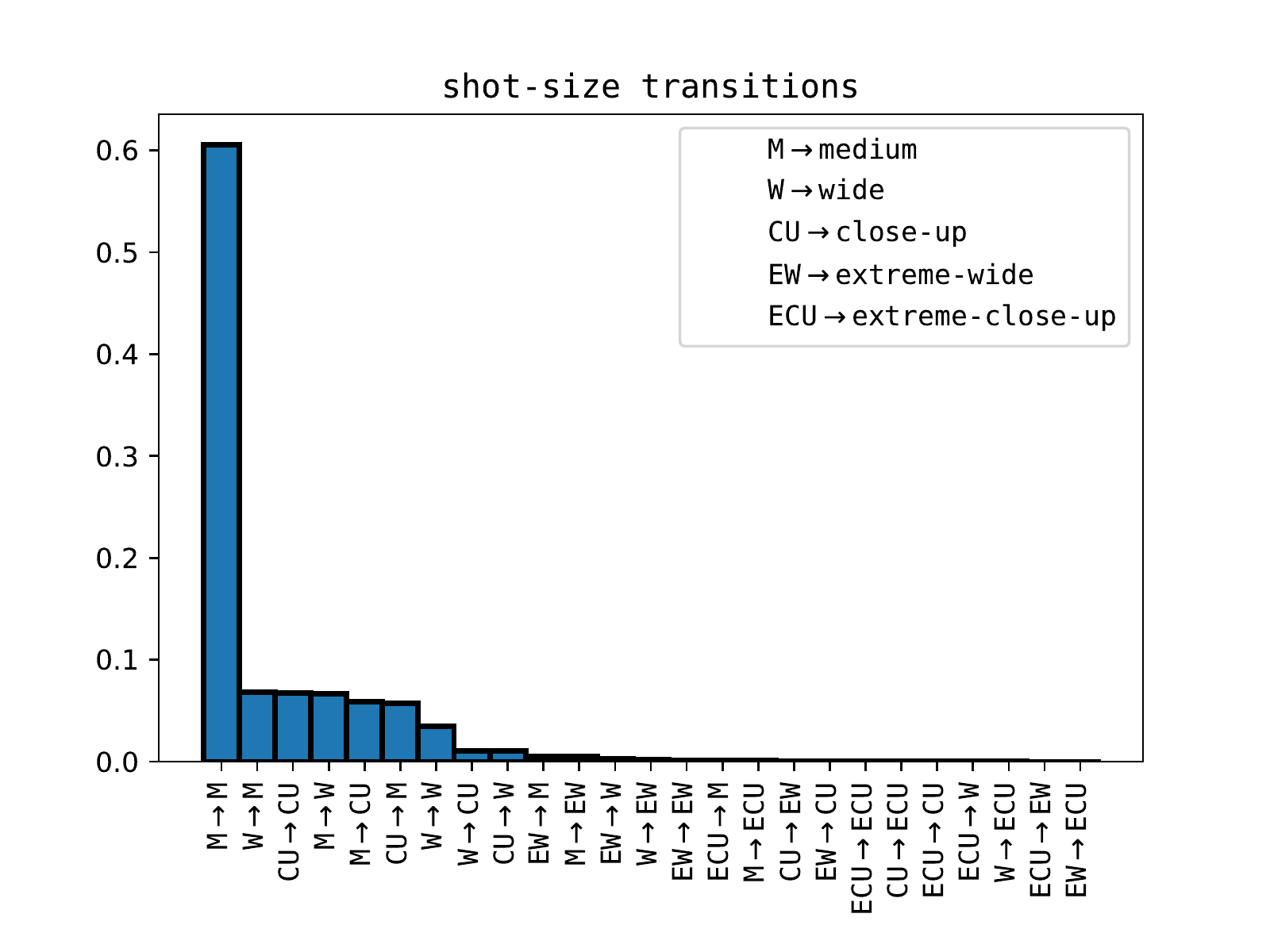}
    \end{subfigure}
    \hfill
    \begin{subfigure}[]{0.48\textwidth}
    \centering
    \includegraphics[width = \textwidth, trim={0.8cm 0.2cm 1.60cm 0.9cm},clip]{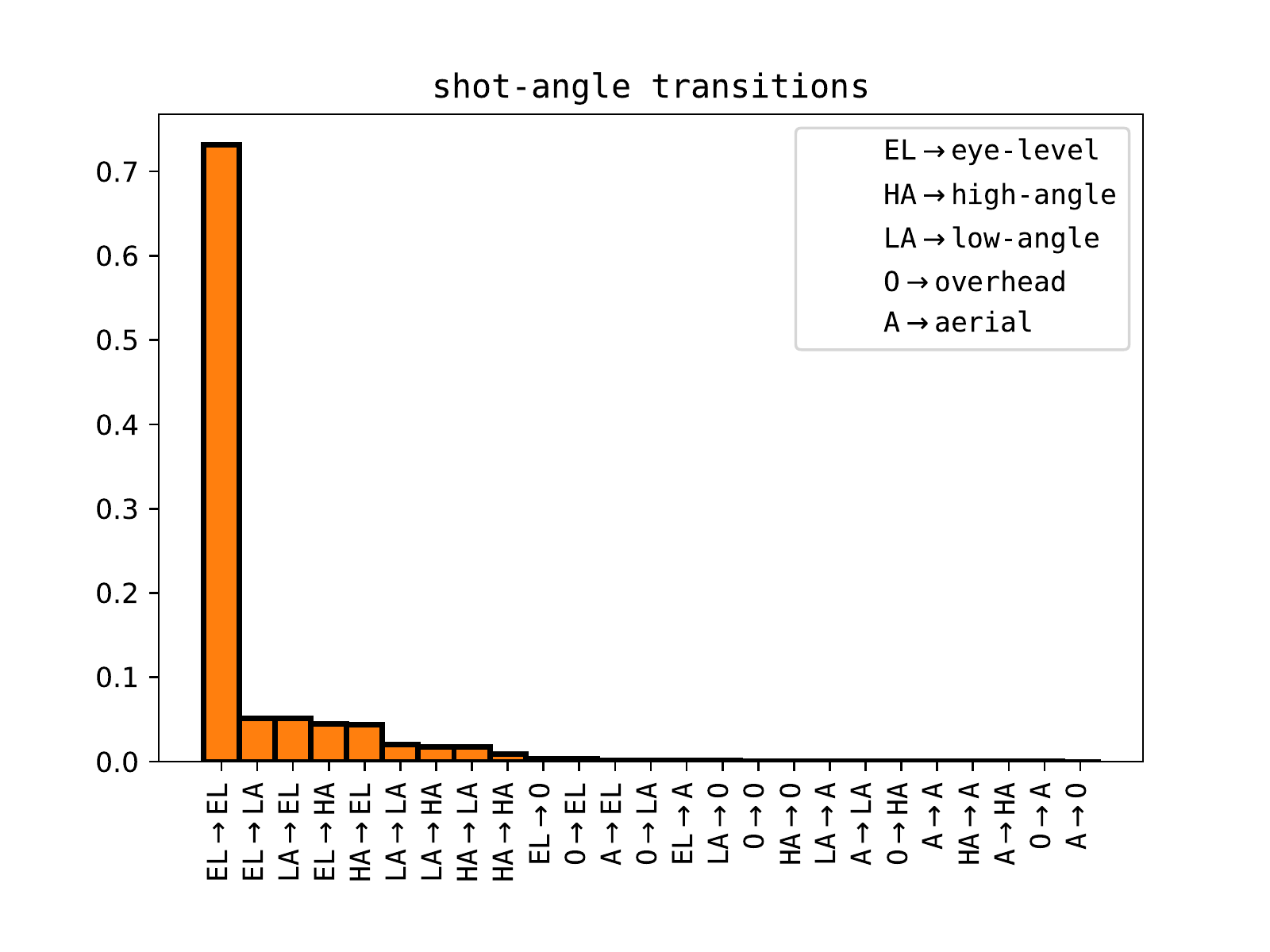}
    \end{subfigure}
    \begin{subfigure}[]{0.48\textwidth}
    \centering
    \includegraphics[width = \textwidth, trim={0.8cm 0.2cm 1.60cm 0.9cm},clip]{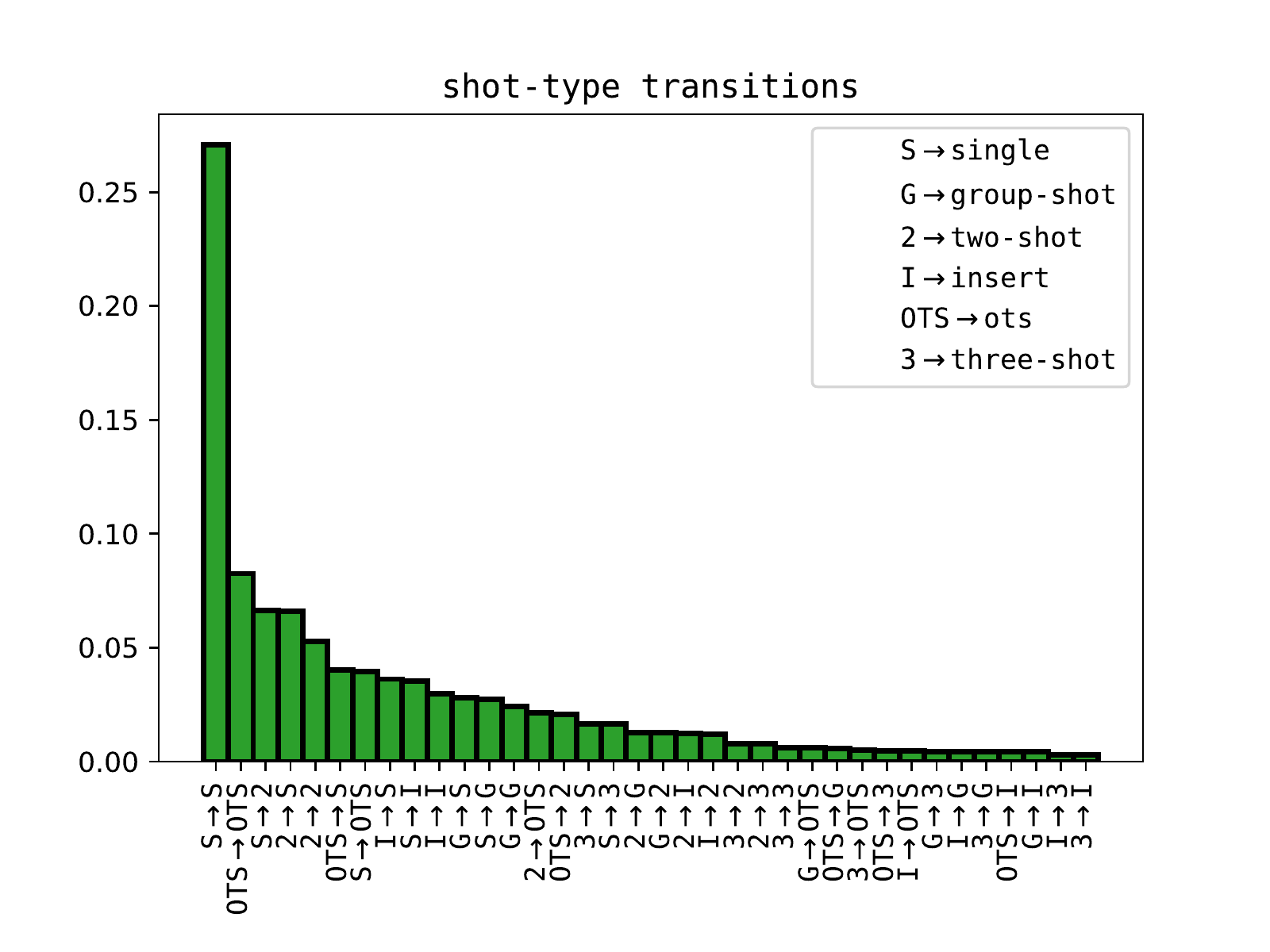}
    \end{subfigure}
    \hfill
    \begin{subfigure}[]{0.48\textwidth}
    \centering
    \includegraphics[width = \textwidth, trim={0.8cm 0.2cm 1.60cm 0.9cm},clip]{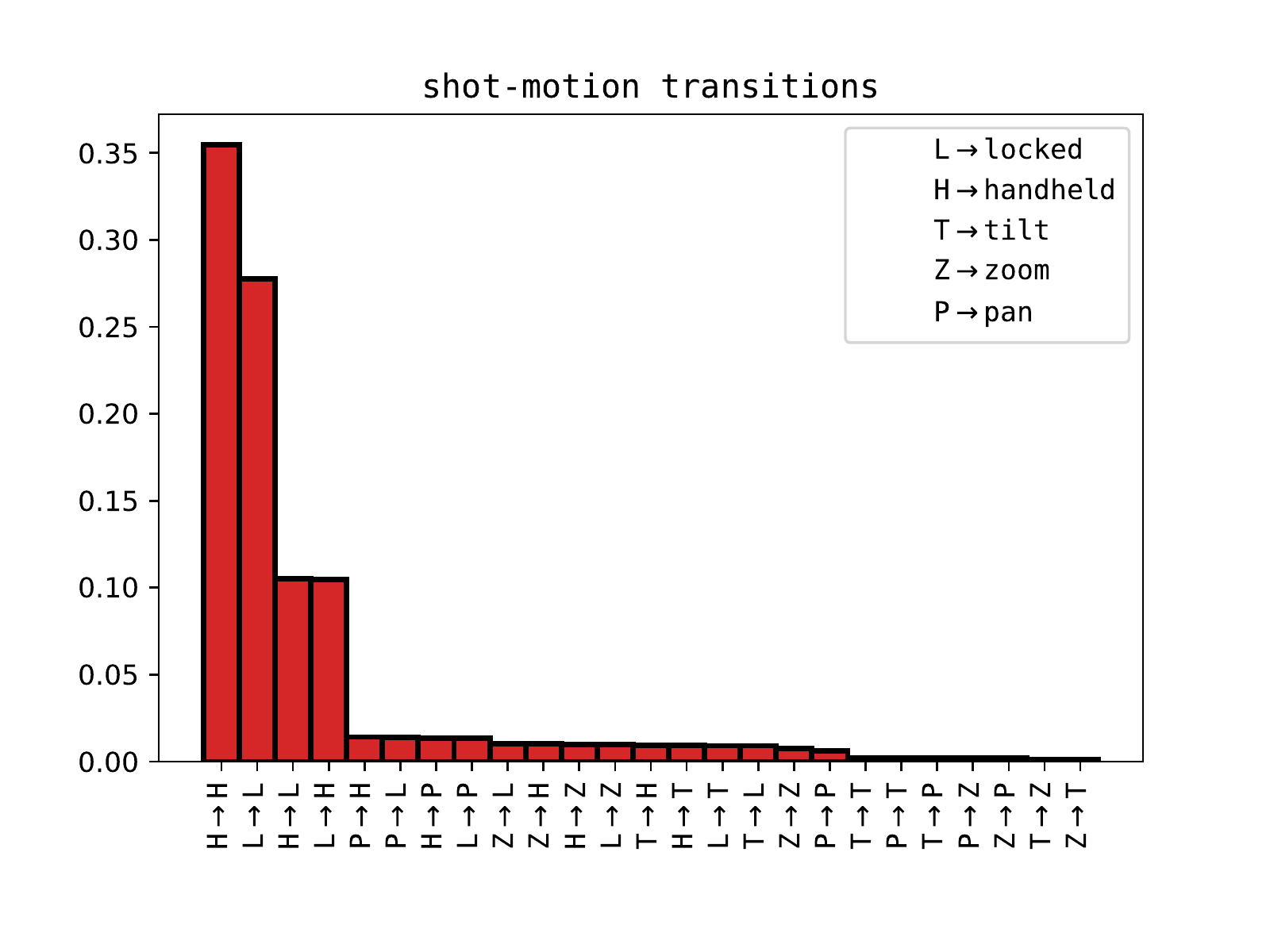}
    \end{subfigure}
    \begin{subfigure}[]{0.48\textwidth}
    \centering
    \includegraphics[width = \textwidth, trim={0.8cm 0.2cm 1.60cm 0.9cm},clip]{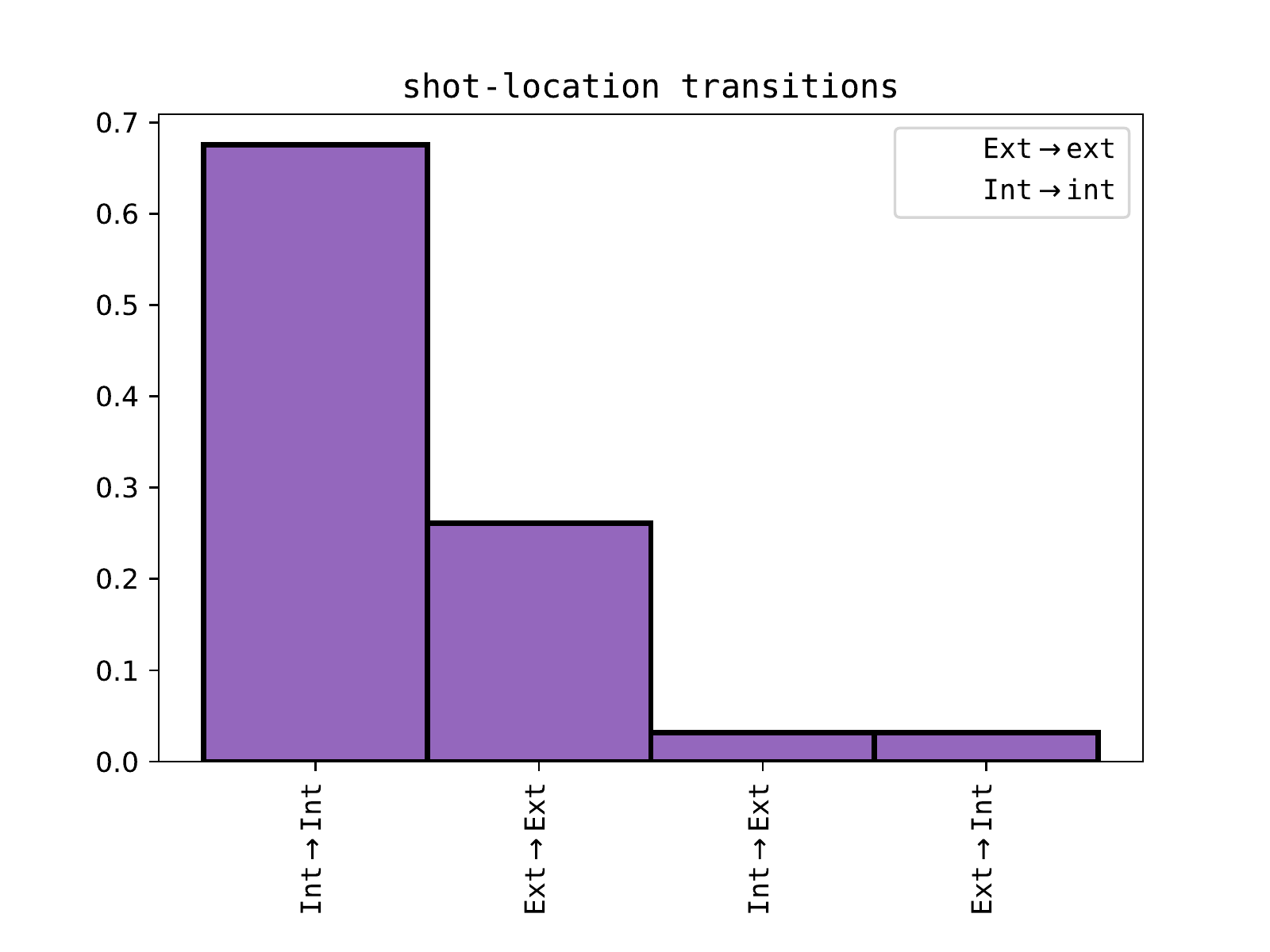}
    \end{subfigure}
    \hfill
    \begin{subfigure}[]{0.48\textwidth}
    \centering
    \includegraphics[width = \textwidth, trim={0.8cm 0.2cm 1.60cm 0.9cm},clip]{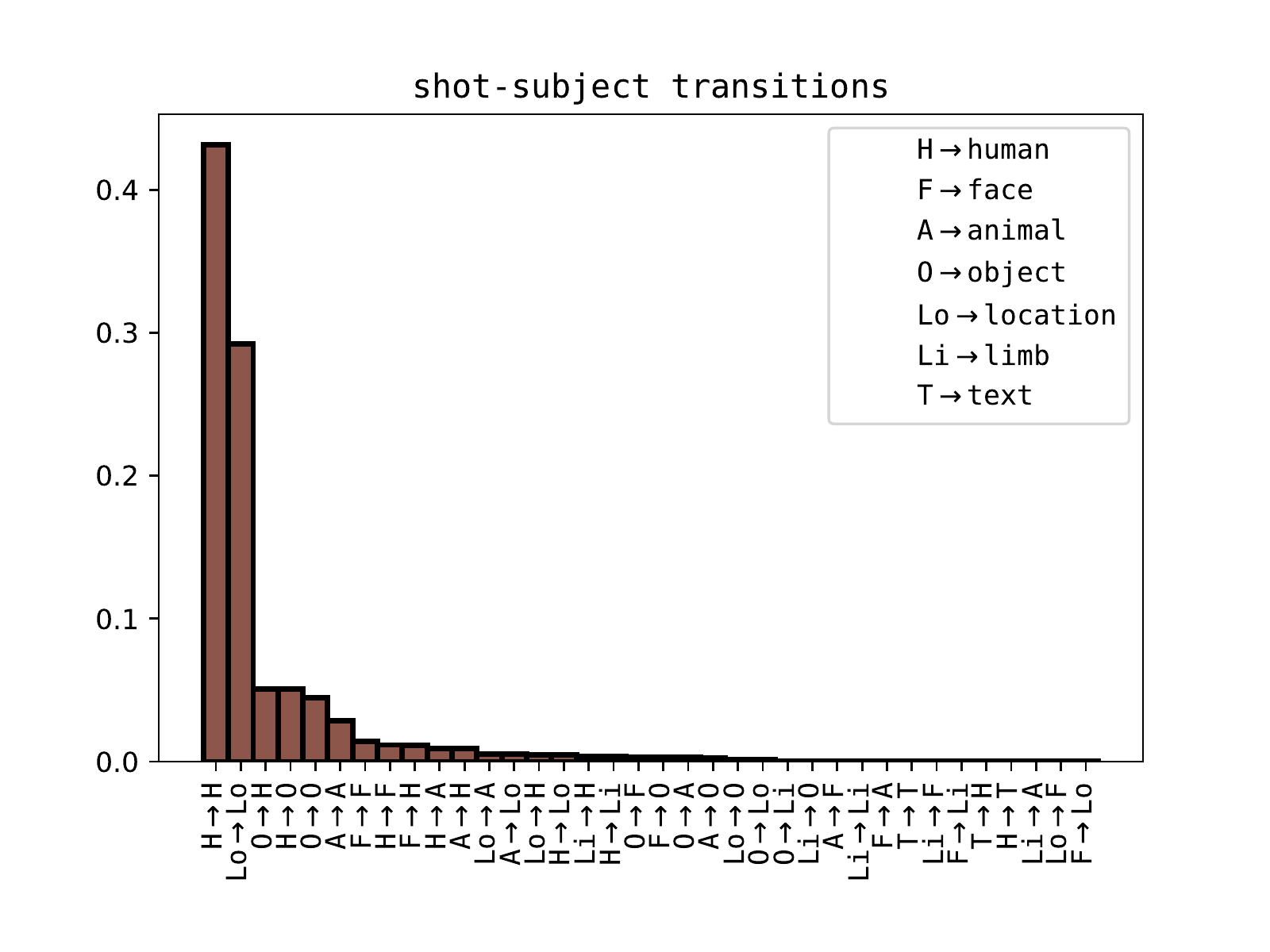}
    \end{subfigure}
    \begin{subfigure}[]{0.48\textwidth}
    \centering
    \includegraphics[width = \textwidth, trim={0.8cm 0.5cm 1.60cm 0.9cm},clip]{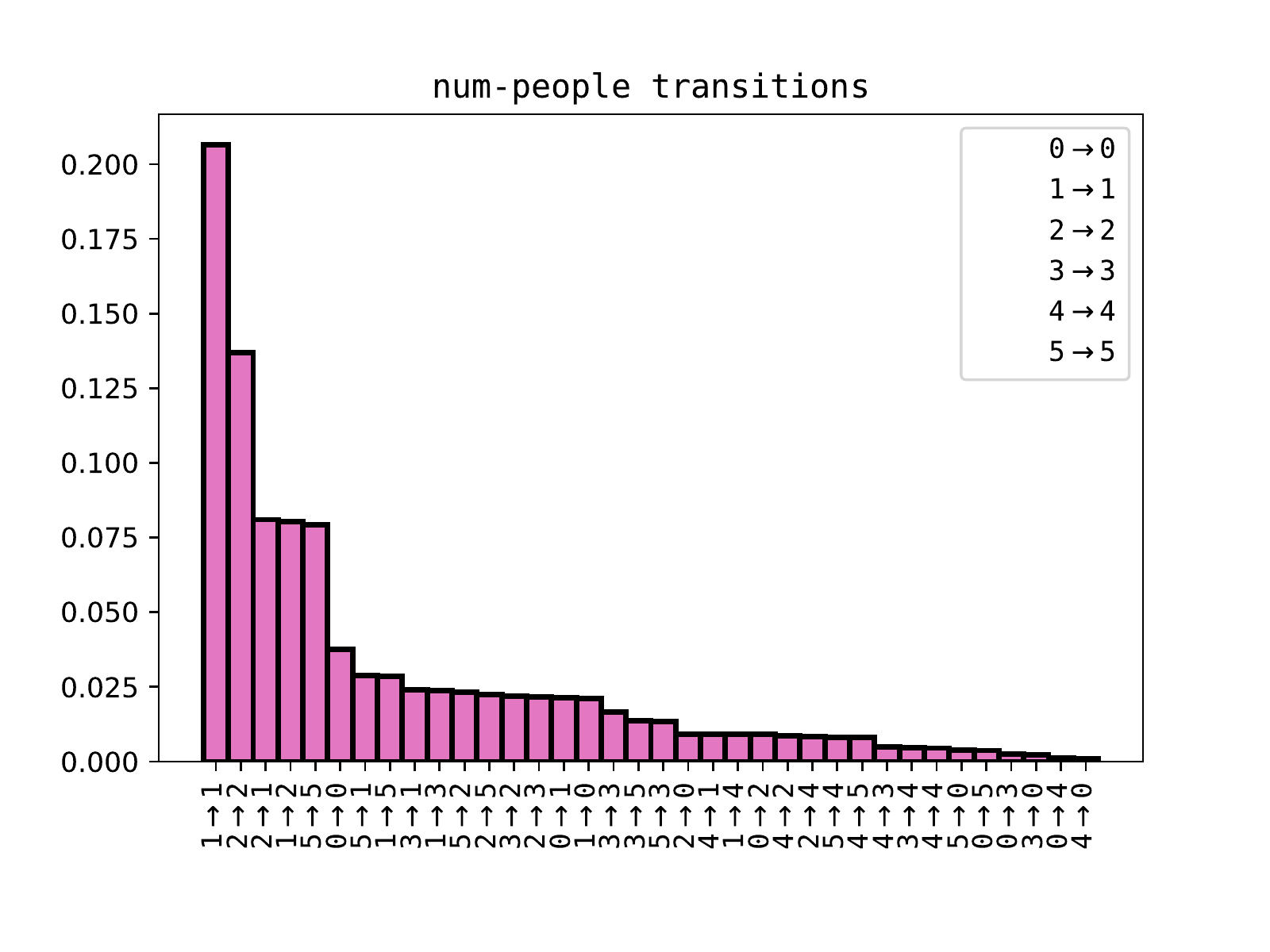}
    \end{subfigure}
    \hfill
    \begin{subfigure}[]{0.48\textwidth}
    \centering
    \includegraphics[width = \textwidth, trim={1cm 0.5cm 1.60cm 0.9cm},clip]{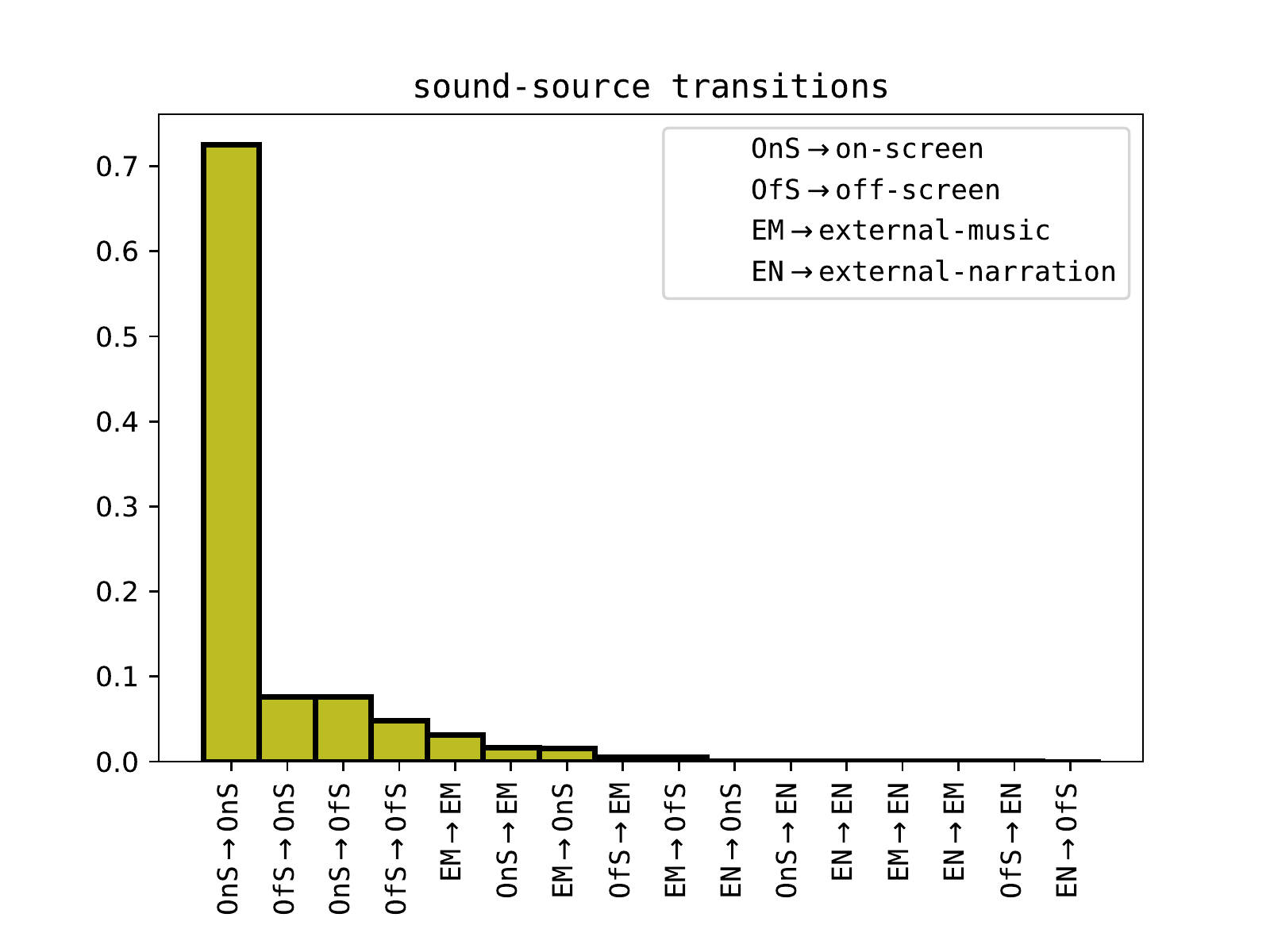}
    \end{subfigure}
    \caption{\small Most frequent shot attribute transitions in \texttt{AVE}. The \texttt{y-axis} indicates the probability of occurrence and the \texttt{x-axis} denotes the transition.}
    \label{fig:qual_trans}
    \vspace{-2mm}
\end{figure*}

Here, we present additional details, results and analyses that could not be included in the main paper due to page-limit constraints. All references and figures in this supplementary file are self-contained. 
\section{Anatomy of Video Editing (AVE): Dataset}
In \Fref{fig:qual_att1}, \Fref{fig:qual_att2} and \Fref{fig:qual_att3}, we plot the class-wise distribution statistics for all shot attributes. We can infer two key things from the figures. First, there is a long tail label distribution problem in many of the shot attributes as discussed in the main paper. Second, we conduct all of our experiments on a very balanced training and evaluation sets. Beyond the shot-level attributes, we further analyze transition patterns between contiguous shots in \texttt{AVE}. In \Fref{fig:qual_trans}, we plot the most frequent shot attribute transitions with respect to their probability of occurrence. Please refer to the \href{https://sites.google.com/view/anatomy-of-video-editing/home}{Project page} for further qualitative analyses on the proposed dataset and benchmark tasks.

\section{Experimental Results and Discussion}
\subsection{Experimental Settings}
\subsubsection{Network Architecture.} We use \texttt{ResNet-101}~\cite{he2016deep} and \texttt{R-3D}~\cite{tran2018closer} as visual backbone networks for image and video inputs, respectively. For feature extraction, we remove the last \texttt{fc} layer from both networks, and obtain a visual feature of size $1024$ and $512$ for \texttt{ResNet-101} and \texttt{R-3D} backbones, respectively. Visual backbone networks are initialized with pretrained weights (\texttt{ResNet-101} - pretrained on ImageNet~\cite{deng2009imagenet} and \texttt{R-3D} - pretrained on Kinetics-400~\cite{kay2017kinetics}) and fine-tuned during training. To extract features from the audio input, we designed \texttt{AudioNet}, which is a feed-forward network with three \texttt{convolutional} and two \texttt{linear} layers. We use \texttt{kernel} sizes of $\{(7 \times 3), (5 \times 3), (5 \times 5)\}$ and \texttt{stride} sizes of $\{(3 \times 1), (3 \times 1), (3 \times 3)\}$ for the 3 \texttt{convolutional} layers, respectively. Each \texttt{convolutional} layer is followed by a \texttt{ReLU} activation layer. After processing the audio input through the \texttt{convolutional} layers, we apply a \texttt{global pooling} layer to obtain a one dimensional audio feature. This feature is further processed using 2 \texttt{linear} layers with a \texttt{ReLU} activation layer in between. The visual and audio features are then concatenated channel-wise to obtain an audio-visual feature. We use the same visual and audio networks for all tasks in the main paper.

 For \textit{shot attributes classification} task, the audio-visual feature is feed into eight classifier networks. We use a network with two \texttt{linear} layers as a classifier network for each attribute. A \texttt{dropout} and a \texttt{ReLU} activation layers are used in between the \texttt{linear} layers. The final \texttt{linear} layer outputs a logit vector with size equal to the number of classes for the respective shot attribute. 

 For \textit{shot sequence ordering} task, the \texttt{feature fusion} network in the late feature fusion baseline (\texttt{Baseline-I}) is a simple network with \texttt{linear}, \texttt{ReLU} activation and \texttt{dropout} layers. The \texttt{feature fusion} network inputs the hierarchically concatenated features and outputs a fused feature representation. The classifier networks in both \texttt{Baseline-I} and \texttt{Baseline-II} (early input fusion) have the same architecture as the classifier in the previous task.
 
 For \textit{next shot selection} task, a recurrent network is used to learn the shot sequence pattern. We use a two-layered (stacked) \texttt{LSTM} module with a hidden size of 128 to obtain \texttt{anchor}, \texttt{positive} and \texttt{negative} embeddings from an input sequence of audio-visual features.

For \textit{missing shot attributes prediction} task, a \texttt{label-to-feature (L2F)} network is used to incorporate the attributes of the input shots along with their audio-visual features. We use a simple 1-layered linear network to transform an attribute vector of size 8 to a feature representation. A \texttt{feature fusion} network (like the one used in shot sequence ordering task) is then used to combine the cross-modal features extracted from the two input shots. The output of the \texttt{feature fusion} network is feed into eight classifiers to predict the attributes of the missing shot. We use the same classifier architecture as the one used in shot attributes classification task.
\vspace{-3mm}

\begin{table}[!t]
\setlength{\tabcolsep}{6pt}
\begin{center}
\caption{\small Train-val-test split statistics for different tasks}
\vspace{-2.5mm}
\label{tbl:supp_dataset}
\mytabular{
\begin{tabular}{l|c|cccc}
\toprule
 & \# of shots & Train & Val & Test & Total \\ \cmidrule(lr) {3-3} \cmidrule(lr) {4-4} \cmidrule(lr) {5-5} \cmidrule(lr) {6-6}

\texttt{Num. of scenes} & - & 3914 & 559 &  1118 & 5591 \\ \midrule
\texttt{Shot attributes classification} & 1 &151053 & 15040  & 30083  & 196176\\
\texttt{Camera setup clustering} &- & - &-& -&-\\
\texttt{Shot sequence ordering} &3&75000& 7500 & 15000 & 97500\\
\texttt{Next shot selection} &9&75000& 7500 & 15000 & 97500\\
\texttt{Missing shot attributes prediction} &3&75000& 7500 & 15000 & 97500\\
\bottomrule
\end{tabular}
}
\end{center}
\vspace{-6.5mm}
\end{table}

\subsubsection{Dataset.} We follow a $\texttt{train-val-test}$ scene split of 70-10-20 in all experiments (see \Tref{tbl:supp_dataset}). As the scenes in the proposed dataset are non-overlapping, the train, validation and test splits are disjoint sets. For \textit{shot attributes classification} task, we use all the shots in the respective scene split for training and evaluation, \ie~151053 training, 15040 validation and 30083 testing shots. For \textit{shot ordering} and \textit{missing shot attributes prediction} tasks, we generate train, validation and test sets by sampling 3 consecutive shots from a scene at a time. As shown in \Tref{tbl:supp_dataset}, we create a total of 97500 shot triplets, where 75000, 7500 and 15000 shot sequences are used for training, validation and testing, respectively. The sequences are randomly shuffled during training for \textit{shot ordering} task, thereby creating an augmented dataset that is significantly larger than the initial samples. For next shot selection task, we sample 9 consecutive shots from a scene at a time. The first 4 shots in the sequence are used as a context. The remaining 5 shots are used to make a candidate list. We generate 75000 training, 7500 validation and 15000 testing shot sequences for conducting experiments. 
\vspace{-3mm}
\subsubsection{Implementation Details.} We use \texttt{ffmpeg} to extract frames of size $1280 \times 720$ from a given shot clip. We crop 50 pixels from the top and bottom corners of each frame to remove the logo of the channel from where the movie scenes are crawled and create consistency across the dataset~\footnote{\href{https://www.youtube.com/user/movieclips}{MovieClips YouTube Channel}}. We then uniformly sample 16 frames and resize each sampled frame to a size of $320 \times 130$ to represent a shot clip as a \texttt{video} input to \texttt{R-3D}. We pick the central frame from the extracted frame sequence of a shot and resize it to $640 \times 260$ to represent a shot clip as a single \texttt{frame} input to \texttt{ResNet-101}. The audio file from a given shot clip is extracted in a `\texttt{.wav}' format using \texttt{ffmpeg}. We then use the \texttt{torchaudio}~\cite{yang2021torchaudio} library to load the `\texttt{.wav}' file as a 2D \texttt{spectrogram} image of size $513 \times 32$, by applying zero padding when necessary. The spectrogram image is then feed into \texttt{AudioNet} to extract audio features. We implement our models in PyTorch~\cite{paszke2019pytorch}. We use SGD optimizer~\cite{ruder2016overview} with a \textit{momentum}, \textit{weight decay} and initial \textit{learning rate}, $0.9$, $1e-4$ and $1e-3$, respectively, for all tasks.

For \textit{shot attributes classification} task, we train our framework for 100 epochs, with the learning rate decaying by 0.1 at 40, 60 and 80 epochs. We use a batch size of 50 during training. We deal with the long tail label distribution problem by adjusting the logits~\cite{menon2020long} of each classifier according to the label frequencies in the respective shot attribute. Note that logit adjustment is used only during training. To scale the cross entropy loss from each classifier for multi-task training, we follow~\cite{liu2019end} and implement dynamic weight averaging technique with a temperature parameter $T = 2$. For single-task training, we simply optimize the cross entropy loss of the classifier.

For \textit{camera setup clustering} task, we experiment with several feature extraction methods. For \texttt{SIFT}~\cite{lowe2004distinctive}, we use the implementation from \texttt{OpenCV}~\footnote{\href{https://docs.opencv.org/4.x/da/df5/tutorial_py_sift_intro.html}{OpenCV SIFT}}. For \texttt{CLIP}~\cite{radford2021learning}, we use the image encoder part of the official pretrained model with `\texttt{ViT-B/32}' backbone. For \texttt{ResNet-101}~\cite{he2016deep} and \texttt{R-3D}~\cite{tran2018closer}, we use the pretrained models (with the last \texttt{fc} layer removed) from PyTorch~\cite{paszke2019pytorch}. For standard clustering algorithms such as \texttt{K-Means}~\cite{lloyd1982least}, Hierarchical Agglomerative Clustering (\texttt{HAC})~\cite{mullner2011modern} and \texttt{OPTICS}~\cite{ankerst1999optics}, we use the \texttt{scikit-learn} implementations~\footnote{\href{hhttps://scikit-learn.org/stable/modules/clustering.html}{Scikit-learn Clustering}}. For \texttt{FINCH}~\cite{sarfraz2019efficient}, we use the official code.

For \textit{shot sequence ordering} task, we train both \texttt{Baseline-I} and \texttt{Baseline-II} for 100 epochs, with the learning rate decaying by 0.1 at 40, 60 and 80 epochs. We used a batch size of 50 during training. We use the cross entropy loss for training both baselines.

For \textit{next shot selection} task, we train our network for 200 epochs, with the learning rate decaying by 0.1 at 100, 150 and 175 epochs. We used a batch size of 100 during training. We use the supervised NT-Xent loss~\cite{chen2020simple,NEURIPS2020_d89a66c7} with a temperature parameter of 0.02 for training. 

For \textit{missing shot attributes prediction} task, we use the same training setting as the shot attributes classification task.
\vspace{-3mm}
\subsection{Experimental Results} 
\subsubsection{Shot Attributes Classification.} In \Tref{tbl:supp_shot_att_1}, we compare the class-wise performance in each attribute for a network trained \textit{with} and \textit{without} taking the long tail label distribution problem into account. Here, we consider a multi-task training setting with \texttt{video + audio} input. As can be seen from \Tref{tbl:supp_shot_att_1}, for attributes with imbalanced label distributions such as \texttt{shot size} and \texttt{shot angle}, nai\"vely trained network performs very well for the dominant classes but extremely poorly for low frequency classes. On the other hand, a network trained with logit adjustment gives a relatively balanced per-class accuracy, and hence a better overall performance. 

\begin{table*}[!t]
\begin{center}
\caption{\small Class-wise performance analysis on shot attributes classification.}
\vspace{-2.5mm}
\label{tbl:supp_shot_att_1}
\setlength{\tabcolsep}{4.5pt}
\renewcommand{\arraystretch}{0.9}

\mytabular{
\begin{tabular}{l|l|cc|cc}
\toprule
& \multicolumn{5}{c}{\textbf{Multi-task training (Video + Audio)}} \\ \midrule
& & \multicolumn{2}{c}{Nai\"ve training} & \multicolumn{2}{c}{Logit adjustment} \\  \cmidrule(lr){3-4} \cmidrule(lr){5-6}
Attribute & & \texttt{Val} & \texttt{Test} & \texttt{Val} & \texttt{Test}  \\ \midrule
\textbf{Shot size} & \texttt{Medium} & 92.7 & 93.0 & 58.0 & 56.0\\
& \texttt{Wide} & 69.4 & 66.2 & 54.7 & 55.0 \\
& \texttt{Close-up} & 19.0 & 22.1 & 67.1 & 65.9\\
& \texttt{Extreme-wide} & 0.0 & 0.0 & 77.5 & 82.8\\
& \texttt{Extreme-close-up} & 0.0 & 0.0 & 61.5  & 65.4\\ \midrule
\rowcolor{cyan}
& \texttt{Average} & 36.2 & 36.4 & 66.8 & 65.0 \\\midrule
\textbf{Shot angle} & \texttt{Eye-level}& 98.1 & 97.6 & 62.3 & 61.7\\
& \texttt{High-angle}& 27.5 & 26.6 & 45.3 & 48.2 \\
&\texttt{Low-angle} & 12.6 & 14.0 & 49.3 & 54.1\\
& \texttt{Overhead}& 0.0 & 0.0 & 43.3 & 18.1\\
&\texttt{Aerial} & 0.0 & 0.0 & 92.9 & 65.4\\ \midrule
\rowcolor{cyan}
&\texttt{Average} & 27.6 & 27.7 & 58.6 & 49.5\\ \midrule

\textbf{Shot type} & \texttt{Single} & 84.6 & 85.9 & 68.1 & 70.7\\
 & \texttt{Group-shot} & 60.7 & 63.5 & 64.1 & 65.1\\
 & \texttt{Two-shot} & 57.9 & 55.3 & 52.3 & 49.7\\
 & \texttt{Insert} & 64.2 & 65.3 & 76.7 & 78.9\\
 & \texttt{OTS} & 68.8 & 70.9 & 75.5& 76.4\\
 & \texttt{Three-shot} & 22.4 & 25.2 & 45.6 & 50.9\\\midrule
 \rowcolor{cyan}
 & \texttt{Average} & 59.8 & 61.0& 63.7 & 65.3 \\\midrule
 
\textbf{Shot motion} & \texttt{Locked} & 79.9 & 79.8 & 82.0 & 82.1 \\
& \texttt{Handheld}& 81.5 & 79.3 & 27.4 & 26.3\\
& \texttt{Tilt}& 0.0 & 0.0 & 40.9 & 38.7 \\
& \texttt{Zoom}& 0.0 & 0.0 & 31.7 & 22.0 \\
& \texttt{Pan}& 0.0 & 0.0 & 41.2 & 47.2 \\\midrule
 \rowcolor{cyan}
 & \texttt{Average} & 32.3 & 31.8 & 44.6 & 43.2 \\\midrule

\textbf{Shot location} & \texttt{Ext} & 73.2 & 67.8 & 84.1 & 81.7\\
 & \texttt{Int} & 92.8 & 94.0 & 83.3& 85.7\\ \midrule
  \rowcolor{cyan}
 & \texttt{Average} & 83.0 & 80.9 & 83.7 & 83.7\\ \midrule
 
\textbf{Shot subject} & \texttt{Human} & 94.9 & 95.3 & 96.0 & 81.2\\
& \texttt{Face} & 93.9 & 94.1 & 90.0& 77.2\\
& \texttt{Animal} & 55.0 & 47.3 & 74.8 & 71.3 \\
& \texttt{Object} & 30.6 & 36.6 & 40.3 & 45.6 \\
& \texttt{Location} & 6.0 & 4.5& 19.3 & 17.6\\
& \texttt{Limb} & 0.0 & 0.0 & 31.2 & 34.0\\
& \texttt{Text} & 0.0 & 0.0 & 0.0 & 0.0 \\ \midrule
\rowcolor{cyan}
& \texttt{Average} & 40.0 & 39.7 & 50.2 & 46.7  \\ \midrule
\textbf{Num. of people} & \texttt{0}& 77.0&74.7 & 90.7 & 88.3 \\
 &\texttt{1} & 81.9 & 82.9 & 73.5 & 76.0 \\
 &\texttt{2} & 72.0 & 72.3 & 62.3 & 62.7\\
 &\texttt{3} & 25.7 & 27.0 & 30.9 & 34.2\\
 &\texttt{4} & 0.0 & 0.1 & 46.1 & 47.8 \\
 &\texttt{5} & 74.0 & 73.3 & 62.1 & 58.5 \\ \midrule
 \rowcolor{cyan}
 &\texttt{Average} & 55.1 & 55.3 & 60.9 & 61.4 \\ \midrule
\textbf{Sound source} & \texttt{On-screen} & 100.0 & 100.0 & 56.7 & 56.3 \\
 & \texttt{Off-screen} & 0.0 & 0.0 & 19.9 & 18.7\\
 & \texttt{External-music} & 0.0 & 0.0 & 41.1 & 46.3\\ 
 & \texttt{External-narration} & 0.0 & 0.0 & 46.1 & 34.3\\ \midrule
 & \texttt{Average} & 25.0 & 25.0 & 41.0 & 38.9 \\ \midrule
 \midrule
  \rowcolor{lime}
\textbf{Average} & &44.9 & 44.7 & \textbf{58.7} & \textbf{56.7} \\
\bottomrule
\end{tabular}
}
\end{center}
\vspace{-6mm}
\end{table*}

\begin{table*}[!t]
\begin{center}
\caption{\small Quantitative analysis on shot attributes classification.}
\vspace{-2.5mm}
\label{tbl:supp_shot_att_2}
\setlength{\tabcolsep}{4.5pt}
\renewcommand{\arraystretch}{0.85}

\mytabular{
\begin{tabular}{l|cccc|cccc|cccc}
\toprule
& \multicolumn{12}{c}{\textbf{Multi-task training}} \\ \midrule
& \multicolumn{4}{c|}{\texttt{Frame}} & \multicolumn{4}{c|}{\texttt{Video}} & \multicolumn{4}{c}{\texttt{Video + Audio}} \\ \cmidrule(lr){2-5}\cmidrule(lr){6-9}\cmidrule(lr){10-13}
&\multicolumn{2}{c}{Nai\"ve} & \multicolumn{2}{c}{Logit adj.} & \multicolumn{2}{c}{Nai\"ve} & \multicolumn{2}{c}{Logit adj.} & \multicolumn{2}{c}{Nai\"ve} & \multicolumn{2}{c}{Logit adj.}\\ \cmidrule(lr){2-3} \cmidrule(lr){4-5} \cmidrule(lr){6-7} \cmidrule(lr){8-9} \cmidrule(lr){10-11} \cmidrule(lr){12-13}
Attribute &\texttt{Val} & \texttt{Test} & \texttt{Val} & \texttt{Test} & \texttt{Val} & \texttt{Test} & \texttt{Val} & \texttt{Test} & \texttt{Val} & \texttt{Test} & \texttt{Val} & \texttt{Test} \\ \midrule

\texttt{Shot size} &38.1 & 37.9 & 62.0 & 66.8 & 35.7 & 35.5 & 67.8 & 66.9 & 36.2 & 36.4 & 66.8 & 65.0 \\
\texttt{Shot angle} &32.7 & 32.6 &63.9 & 55.9 &25.8 & 25.8 & 62.2 & 53.2 & 27.6 & 27.7 & 58.6 & 49.5 \\
\texttt{Shot type} &62.0 &63.2 &64.3& 64.7&59.5 & 60.8 & 63.9 & 64.9 & 59.8 & 61.0 & 63.7 & 65.3 \\
\texttt{Shot motion} &26.4 & 26.0 & 31.7& 33.5&32.1 & 31.7 & 42.8 & 42.7 &32.3 & 31.8 & 44.6 & 43.2 \\
\texttt{Shot location} & 82.6 & 80.0 & 84.0 & 82.1 &82.9 & 81.9 & 84.4 & 83.3 &83.0 & 80.9 & 83.7 & 83.7\\
\texttt{Shot subject} &42.4& 41.2 & 51.0 & 46.8 &40.0 & 39.8 & 50.8 & 47.4 &40.0 & 39.7 & 50.2 & 46.7\\
\texttt{Num. of people} & 57.9 & 56.8 & 61.6 &60.2 &55.0 & 55.1 & 61.3 & 61.2&55.1 & 55.3 & 60.9 & 61.4\\
\texttt{Sound source} & 25.0 &25.0 &31.3 &32.0 & 25.0 &25.0 & 34.4 & 32.6 & 25.0 & 25.0 & 41.0 & 38.9\\ \midrule
\texttt{Average} & 45.9 & 45.3 & \textbf{56.2} & \textbf{55.2} &44.5 & 44.4 & \textbf{58.4} & \textbf{56.5} & 44.9 & 44.7 & \textbf{58.7} & \textbf{56.7} \\
\bottomrule
\end{tabular}
}
\end{center}
\vspace{-6mm}
\end{table*}

In \Tref{tbl:supp_shot_att_2}, we summarize the results of using different input representations for shot attributes classification task in a multi-task training setting. It can be inferred from \Tref{tbl:supp_shot_att_2} that using a single \texttt{frame} to represent a shot generally results in a lower performance compared to using \texttt{video} and \texttt{video + audio}. However, it is worth noticing that, for attributes such as \texttt{shot size} and \texttt{shot angle} which, in essence, does not require temporal or audio information, using a \texttt{frame} representation outperforms other types of inputs. On the other hand, for attributes such as \texttt{shot motion} and \texttt{sound source} which are closely associated with temporal and audio contexts, respectively, using only \texttt{frame} as an input gives a significantly worse performance.
\vspace{-3mm}

\subsubsection{Missing Shot Attributes Prediction.} In \Tref{tbl:supp_missing_shot_1}, we present the results on four shot attributes, \ie~\texttt{shot location}, \texttt{shot subject}, \texttt{num. of people} and \texttt{sound source}, for a model trained in a multi-task setting. This is in continuation of the results from Table 7 in the main paper. As can be seen from \Tref{tbl:supp_missing_shot_1}, the proposed model outperforms the nai\"ve \texttt{dominant label} prediction baseline by a large margin. It can also be inferred that incorporating the attributes of the input shots along with other representations consistently improves model accuracy across all attributes.

\begin{table*}[!b]
\begin{center}
\caption{\small Quantitative analysis on missing shot attributes prediction.}
\vspace{-2.5mm}
\label{tbl:supp_missing_shot_1}
\setlength{\tabcolsep}{4.5pt}
\renewcommand{\arraystretch}{0.85}

\mytabular{
\begin{tabular}{l|cc|cc|cc|cc}
\toprule
& \multicolumn{2}{c|}{\textbf{Shot location}} & \multicolumn{2}{c|}{\textbf{Shot subject}} & \multicolumn{2}{c|}{\textbf{Num. of people}} & \multicolumn{2}{c}{\textbf{Sound source}}\\ \cmidrule(lr){2-3} \cmidrule(lr){4-5} \cmidrule(lr){6-7} \cmidrule(lr){8-9}
Method & \texttt{Val} & \texttt{Test} & \texttt{Val} & \texttt{Test} &  \texttt{Val} & \texttt{Test} & \texttt{Val} & \texttt{Test} \\ \midrule
\texttt{Dominant label} & 50.0 &50.0 & 14.3& 14.3& 16.7&16.7&25.0 & 25.0 \\\midrule
\texttt{Frame} & 73.0 & 70.7 & 27.1 & 23.0 & 31.3 & 26.8 & 31.2 & 31.6\\
\texttt{Frame + Attributes} &94.6&94.4&43.4&43.1&37.9&37.4& 34.8 & 37.1 \\  \midrule
\texttt{Video} &83.8&82.2&30.3&27.4&36.6&35.4&28.8&32.6 \\
\texttt{Video + Attributes} &93.9&92.8&44.4&43.9&38.4&37.4&35.7&33.8 \\  \midrule
\texttt{Video + Audio} & 85.8 & 83.4 & 31.0 & 28.7 & 37.3 & 36.3 & 29.7 & 33.3 \\
\texttt{Video + Audio + Attributes} &\textbf{ 95.0} & \textbf{94.4} & \textbf{45.5} & \textbf{45.0} & \textbf{38.9} & \textbf{38.4} & \textbf{40.3}&\textbf{41.2} \\
\bottomrule
\end{tabular}
}
\end{center}
\vspace{-5mm}
\end{table*}

As shown in \Tref{tbl:supp_missing_shot_2}, the multi-task training setup leads to an unbalanced performance when using \texttt{frame} as an input, \ie~the performance gap between \texttt{shot size} and other attributes is notably large in comparison with using other input representations (refer to Table 7 in the main paper). This is mainly because the model overfitted to the \texttt{shot size} attribute for this particular input setup. To verify this hypothesis,  we train our model in a single-task setting for each attribute. As can be inferred from \Tref{tbl:supp_missing_shot_2}, the single-task setting gives a relatively balanced performance across attributes.

\begin{table*}[!t]
\begin{center}
\caption{\small Multi-task vs. single-task analysis on missing shot attributes prediction.}
\vspace{-2.5mm}
\label{tbl:supp_missing_shot_2}
\setlength{\tabcolsep}{4.5pt}
\renewcommand{\arraystretch}{0.85}

\mytabular{
\begin{tabular}{l|l|cc|cc|cc|cc}
\toprule
& & \multicolumn{2}{c|}{\textbf{Shot size}} & \multicolumn{2}{c|}{\textbf{Shot angle}} & \multicolumn{2}{c|}{\textbf{Shot type}} & \multicolumn{2}{c}{\textbf{Shot motion}}\\ \cmidrule(lr){3-4} \cmidrule(lr){5-6} \cmidrule(lr){7-8} \cmidrule(lr){9-10}
Setting & Method & \texttt{Val} & \texttt{Test} & \texttt{Val} & \texttt{Test} &  \texttt{Val} & \texttt{Test} & \texttt{Val} & \texttt{Test} \\ \midrule

&\texttt{Dominant label} & 20.0 & 20.0 & 20.0 & 20.0 & 16.7 & 16.7 & 20.0 & 20.0\\ \midrule
\texttt{Multi-task} & \texttt{Frame} & 40.9 & 32.4 & 22.6 & 30.5 & 26.6 & 26.1 & 25.0 & 25.8 \\
& \texttt{Frame + Attributes} & 47.8 & 44.6 & 28.5 & 34.1 & 32.0 & 34.5 & 31.0 & 31.8\\  \midrule
\texttt{Single-task} & \texttt{Frame} & 29.4 & 32.2 & 25.6 & 28.4 & 26.5 & 25.2 & 27.2 & 27.7\\
& \texttt{Frame + Attributes} & 34.3 & 36.6 & 30.3 & 32.9 &36.0 & 34.6 & 29.4  & 30.2 \\ 
\bottomrule
\end{tabular}
}
\end{center}
\vspace{-5mm}
\end{table*}

\subsection{Discussion} 
Here, we discuss the concerns raised by anonymous reviewers regarding shot-attributes classification task. The additional results that addressed the concerns are presented in \Tref{tbl:rebuttal}. Experiments are conducted in a multi-task setting applying the logit adjustment~\cite{menon2020long} technique.
\vspace{-3mm}
\paragraph{Why not use SOTA methods such as CLIP?} We experimented with using pretrained CLIP's visual encoder as a backbone network for shot attributes classification task, however, we observed an inferior performance compared to using ResNet-101 (see \textbf{CLIP} column in \Tref{tbl:rebuttal}).
\vspace{-3mm}
\paragraph{Why use vector concatenation for feature fusion?} Because it is simple. We also experimented with weighted combination of visual and audio features as a fusion mechanism, however, we did not observe any significant improvement in performance (see \textbf{Weight} column in \Tref{tbl:rebuttal})
\vspace{-3mm}
\paragraph{Did you try freezing the backbones?}
We did. However, we opted for fine-tuning the backbone networks along with the classifiers because it resulted in a much better performance (see \textbf{Freeze} column in \Tref{tbl:rebuttal}).

\begin{table}[!t]
\begin{center}
\caption{\small Additional results for shot attributes classification task}
\vspace{-3mm}
\label{tbl:rebuttal}
\setlength{\tabcolsep}{4.5pt}
\renewcommand{\arraystretch}{0.85}

\mytabular{
\begin{tabular}{l|cccc|cccccc}
\toprule
&\multicolumn{4}{c|}{\texttt{Frame}} & \multicolumn{6}{c}{\texttt{Audio + Video}} \\\cmidrule{2-5}
\cmidrule{6-11}
&\multicolumn{2}{c}{\textbf{CLIP}} & \multicolumn{2}{c|}{\textbf{ResNet-101}} & \multicolumn{2}{c}{\textbf{Weight}}  & \multicolumn{2}{c}{\textbf{Freeze}} & \multicolumn{2}{c}{\textbf{Baseline}} \\\cmidrule{2-3} \cmidrule{4-5} \cmidrule{6-7} \cmidrule{8-9} \cmidrule{10-11}
Attribute & \texttt{Val} & \texttt{Test} & \texttt{Val} & \texttt{Test} & \texttt{Val} & \texttt{Test} & \texttt{Val} & \texttt{Test} & \texttt{Val} & \texttt{Test}\\ \midrule

\texttt{Shot size} & 52.9 & 51.3 & 62.0 & 66.8 & 66.9 & 66.0 & 51.9 & 50.4 & 66.8 & 65.0\\
\texttt{Shot angle} & 54.3 & 54.9 &  63.9 & 55.9 & 59.0 & 50.2 & 54.2 & 54.3 & 58.6 & 49.5\\
\texttt{Shot type} & 56.0 & 58.0 & 64.3& 64.7 & 62.4 & 64.6 & 49.4 & 51.2 & 63.7 & 65.3\\
\texttt{Shot motion} & 36.6 & 37.6 & 31.7& 33.5 & 44.7 & 43.2 & 35.6 & 35.2 & 44.6 & 43.2\\
\texttt{Shot location} & 82.1 & 81.0& 84.0 & 82.1 & 83.8 & 83.7 & 84.6 & 85.0 & 83.7 & 83.7\\
\texttt{Shot subject} & 45.2 & 42.9 & 51.0 & 46.8 & 48.8 & 45.8 & 47.6 & 43.9 & 50.2 & 46.7\\
\texttt{Num. of people} & 56.9 & 57.2 & 61.6 &60.2 & 60.0 & 61.4& 53.3 & 52.6 & 60.9 & 61.4 \\
\texttt{Sound source} & 44.1 & 43.4& 31.3 &32.0 & 43.1 & 40.1 & 41.9 & 40.9 & 41.0 & 38.9\\ \midrule
\texttt{Average} & 53.5 & 53.1 & \textbf{56.2} & \textbf{55.2} & 58.6 & \textbf{56.9} & 52.3 & 51.7 & \textbf{58.7} & 56.7\\
\bottomrule
\end{tabular}
}
\end{center}
\vspace{-8mm}
\end{table}
\clearpage
%
%

\bibliographystyle{splncs04}
\bibliography{egbib}